\documentclass[a4paper,10pt]{academic_template}
\usepackage{acap_preprint}

\usepackage{amsmath,amssymb,mathtools}
\usepackage{graphicx,booktabs,array,tabularx,multirow,longtable,placeins}
\usepackage{xcolor}
\usepackage{tikz}
\usetikzlibrary{arrows.meta,positioning,fit,backgrounds}
\usepackage{enumitem}
\usepackage{hyperref}
\usepackage{url}
\hypersetup{pdftitle={ACAP: Asynchronous Coding Agents for Efficient Robot Control},pdfauthor={Zexin He, Rui Li, Yushi Lan, Chengkai Jin, Zhi Hou, Xingang Pan}}

\title{ACAP: Asynchronous Coding Agents\\for Efficient Robot Control}

\author{%
\textbf{Zexin He}\textsuperscript{1}\quad \textbf{Rui Li}\textsuperscript{1}\quad \textbf{Yushi Lan}\textsuperscript{2}\quad \textbf{Chengkai Jin}\textsuperscript{1}\quad \textbf{Zhi Hou}\textsuperscript{3}\quad \textbf{Xingang Pan}\textsuperscript{1,3}\\[0.4em]
{\small \textsuperscript{1}S-Lab, Nanyang Technological University\quad \textsuperscript{2}University of Oxford\quad \textsuperscript{3}ACE Robotics}%
}

\newcommand{\method}{\textup{\textsc{ACAP}}}
\newcommand{\lead}{\mathrm{L}}
\newcommand{\member}{\mathrm{M}}
\newcommand{\owner}{\omega}

\makeatletter
\let\inputrows\@@input
\makeatother
\newcolumntype{Y}{>{\raggedright\arraybackslash}X}
\crefname{figure}{Figure}{Figures}
\crefname{table}{Table}{Tables}
\crefname{section}{Section}{Sections}
\crefname{appendix}{Appendix}{Appendices}
\Crefname{appendix}{Appendix}{Appendices}
\crefname{algorithm}{Algorithm}{Algorithms}
\Crefname{algorithm}{Algorithm}{Algorithms}
\creflabelformat{equation}{#2#1#3}
\AtBeginDocument{%
}

\begin{document}
\maketitle
\begingroup
\setlength{\intextsep}{8pt}
\begin{figure}[!h]
\centering
\includegraphics[width=\linewidth]{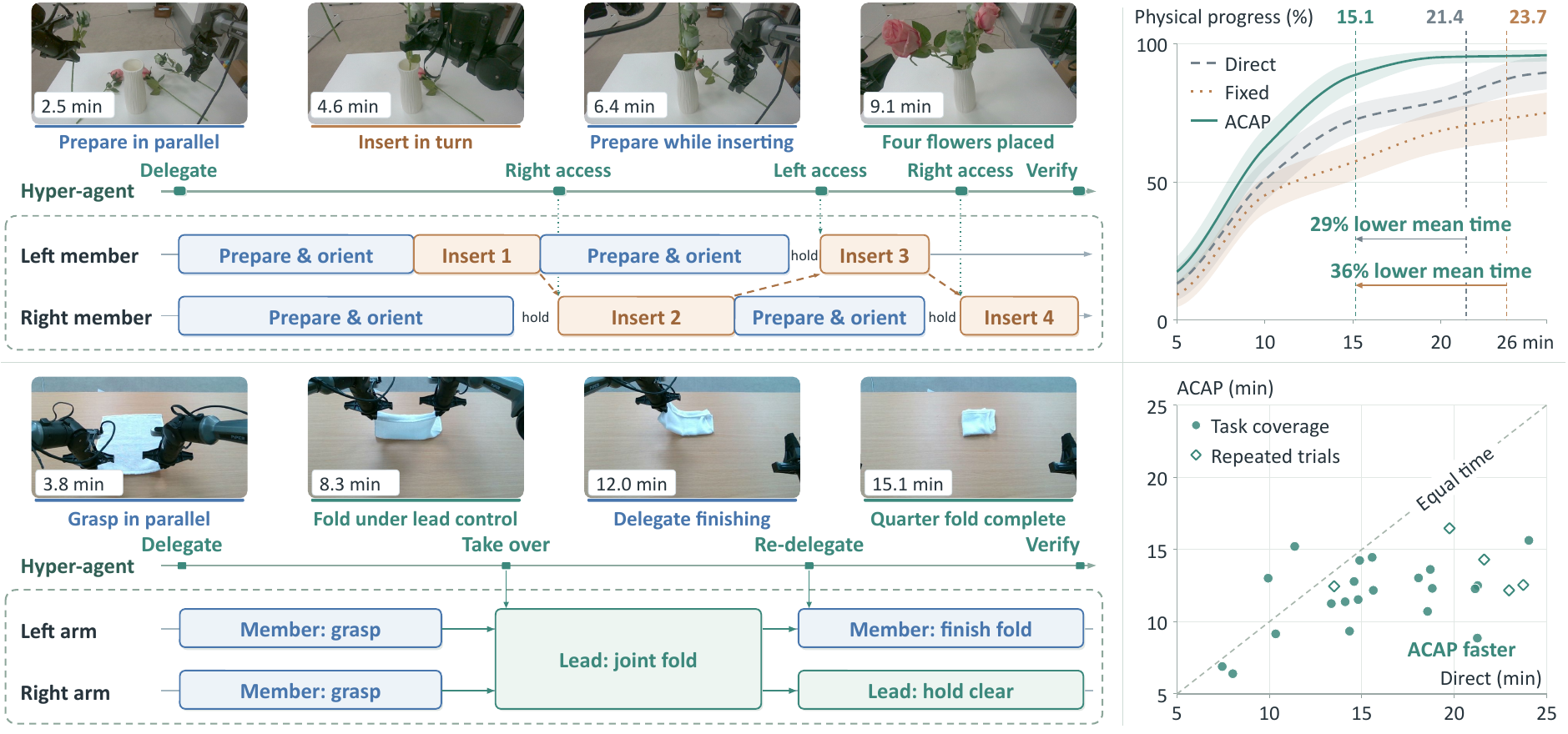}
\caption[Asynchronous reasoning with agent-directed coordination.]{\textbf{Asynchronous reasoning with agent-directed coordination.} Left: Schematic timelines show parallel preparation with ordered vase access, and lead takeover followed by delegated finishing for towel folding. Top right: Mean physical progress and time cost across six tasks, with five trials per policy. Bottom right: Task-level time comparisons between Direct and ACAP in the task-coverage and repeated-trial studies.}
\label{fig:teaser}
\end{figure}
\FloatBarrier
\endgroup

\phantomsection
\addcontentsline{toc}{section}{Abstract}
\begin{abstract}
Coding agents can construct and revise robot behavior during execution, but using one agent for the whole robot can delay otherwise independent work. We introduce Asynchronous Coding Agents as Policy (\method), an asynchronous framework that organizes \emph{one robot as a team of persistent agents}. Each member reasons and acts for a configurable control group, such as an arm with its gripper or the torso and chassis, enabling asynchronous reasoning and execution. However, different robot components often have physical dependencies that require coordination. To address this challenge, \method\ introduces a hyper-agent that assigns tasks, coordinates dependencies, and dynamically allocates control between the lead and member agents. This allows independent work to progress concurrently while supporting tightly coupled manipulation when necessary. Across 72 real-world rollouts covering 25 task/platform settings, \method\ reduces task-balanced time cost by 28.2\% compared with a single-agent Direct Policy, with improvements across Fully Parallel, Parallel-Sequential, and Collaborative tasks. Across all eight Collaborative settings, \method\ also achieves lower mean time cost than Fixed Delegation, where only member agents can control their assigned groups. An additional 90-rollout study shows lower mean 30-minute-capped time cost than Direct on all six tasks. We further demonstrate mobile manipulation using the same member interface for chassis and torso control. We present the framework and a real-world rollout dataset containing observations, generated programs, actions, and feedback to support research on how coding agents reason and act in the physical world.
\par\vspace{7pt}
\noindent\textbf{Project page:}~\url{https://projects.zxhezexin.com/acap}
\end{abstract}
\clearpage

\begingroup
\setlength{\parskip}{0pt}
\tableofcontents
\endgroup
\clearpage

\begingroup
\captionsetup{skip=6pt}
\setlength{\textfloatsep}{10pt plus 2pt minus 2pt}
\setlength{\floatsep}{10pt plus 2pt minus 2pt}
\setlength{\intextsep}{10pt plus 2pt minus 2pt}
\section{Introduction}
\label{sec:intro}

Recent coding-agent policies can inspect a scene, construct numerical models, and write and revise robot-control programs during execution~\citep{agp,capx}. They extend the executable-policy interface of Code as Policies~\citep{cap} into sustained interaction with physical feedback. While learning from demonstrations produces increasingly capable visuomotor policies~\citep{diffusion,openvla,pi0}, coding agents can construct task-specific procedures. This flexibility raises a practical question: \emph{how can we organize coding agents to achieve efficient robotic manipulation?}

Reasoning takes time. In a typical coding-agent policy, a single agent is responsible for the entire robot. Although it can issue concurrent motion commands, decisions about different robot components still pass through the same reasoning process. For example, diagnosing a failed grasp with one arm may delay an independent action with the other. This suggests that distributing reasoning across different robot components could improve task efficiency.

We introduce Asynchronous Coding Agents as Policy (\method), an asynchronous framework that \emph{organizes a single robot as a team of persistent coding agents}. Each agent is assigned to a control group, such as an arm with its gripper or a body group comprising the torso and chassis. Each member observes, reasons, and acts within its scope while other members continue their work, enabling parallel reasoning and execution. Member agents can use lightweight reasoning configurations for local tasks, while a more capable coordinating agent reasons about the overall task.

However, independent reasoning does not imply independent physical actions. The agents share one physical robot, where one group's actions can affect another's reach, support, or clearance. A straightforward approach assigns each control group permanently to a member agent. While this can work well for independent tasks, tightly coupled manipulation, such as folding cloth or opening a bottle, requires close coordination between agents. Repeated communication and synchronization can introduce substantial overhead, while one agent's recovery actions may change the object state and invalidate another's plan. The challenge is therefore to coordinate agents efficiently as physical dependencies change.

To address this challenge, \method\ introduces a \emph{hyper-agent} that orchestrates the member agents. Unlike a fixed delegation scheme, the hyper-agent decides not only what tasks to delegate, but also when the lead should take direct control of multiple robot components. It uses task progress and execution feedback to coordinate dependencies and dynamically allocate control between the lead and member agents. For example, in four-flower insertion (\cref{fig:teaser}), the hyper-agent coordinates access to the shared vase while member agents prepare flowers independently. For a shared towel-folding action, the lead can instead directly control both arms while members continue their analysis. The division of control can therefore adapt to the physical dependencies of the task.

We evaluate \method\ on real-world manipulation tasks, including Fully Parallel, Parallel-Sequential, and Collaborative tasks. Across 72 real-world rollouts covering 25 task/platform settings, ACAP reduces the task-balanced time cost by 28.2\% compared with a single-agent Direct Policy, with improvements across all three task categories. To evaluate agentic coordination, we compare ACAP with Fixed Delegation, where only members control their assigned groups. ACAP achieves lower mean time cost across all eight evaluated Collaborative settings, supporting the benefits of flexible coordination for collaborative manipulation. An additional 90-rollout repeated-trial study shows lower mean 30-minute-capped time cost than Direct on all six tasks. We also demonstrate ACAP in mobile manipulation. Furthermore, we collect a real-world rollout dataset containing observations, generated programs, agent interactions, actions, and physical feedback to support future research on embodied coding agents.

Our contributions are threefold:
\begin{enumerate}[leftmargin=*,itemsep=2pt,topsep=3pt,parsep=0pt]
	\item \textbf{Asynchronous agentic policy framework.} We introduce \method, which organizes a single robot as a team of persistent coding agents assigned to configurable control groups, enabling asynchronous reasoning and execution across different robot components.
	\item \textbf{Agentic coordination.} We introduce a hyper-agent that coordinates task dependencies and dynamically manages control allocation between the lead and member agents, supporting both independent and tightly coupled manipulation.
	\item \textbf{Real-world evaluation and rollout dataset.} We evaluate the efficiency and coordination capabilities of ACAP across diverse real-world manipulation tasks. We also present a rollout dataset containing observations, generated programs, agent interactions, and physical execution feedback.
\end{enumerate}

\section{Related work}
\label{sec:related}

\textbf{Coding agents as robot policies.}
Language models ground plans in skills and feedback~\citep{saycan,inner,react}. Code as Policies and related systems express robot behavior as programs~\citep{cap,progprompt,roboscript,robocodex}, while VoxPoser and ReKep compute spatial maps and constraints~\citep{voxposer,rekep}. CodeAct~\citep{codeact} and interactive robot policies~\citep{agp,capx} use code as an agent's action interface. Recent GPT and Claude studies examine real-world robot control~\citep{gptembodied,clauderobotics,openrua} and policy development in simulation~\citep{agenticpusht}. Harness VLA composes fixed analytic primitives with a frozen VLA~\citep{harnessvla}. \method\ lets coding agents revise models, motion programs and recovery code, and distributes this work across persistent agents.

\noindent\textbf{Organizing agents within a robot.}
Agent frameworks organize roles and communication~\citep{autogen,metagpt,agentverse,gptswarm}, including recursive task decomposition~\citep{adapt}. Robotics systems use functional hierarchies~\citep{malmm,hirobot} or distribute work across embodiments~\citep{roco,smartllm,coela,combo}. \method\ assigns agents to control groups within one robot, each observing, reasoning and acting. This permits lighter local reasoning, related to cost-aware model selection~\citep{frugalgpt,routellm}, while multi-agent failure analyses motivate examining coordination costs~\citep{mast}.

\noindent\textbf{Asynchronous reasoning and execution.}
Embodied First Exam includes a real-time setting where simulation continues during reasoning~\citep{embodiedexam}. CaP-X generates programs in parallel for central synthesis~\citep{capx}, LLMCompiler schedules parallel calls by dependency~\citep{llmcompiler}, and Real-Time Chunking overlaps action generation and execution~\citep{rtc}. EMOS combines group discussion with asynchronous execution across robots~\citep{emos}. \method\ maintains persistent member policies whose programs and local knowledge persist across assignments and control transfers. Members can thus reason and recover concurrently while dependent physical actions wait.

\noindent\textbf{Coordinating physical dependencies.}
Bimanual taxonomies and learned role assignment describe spatial and temporal coupling and complementary arm roles~\citep{taxonomy,stabilize}. LABOR selects coordination patterns, DAG-Plan allocates dependency-graph tasks, and LLM+MAP generates symbolic partial-order plans~\citep{labor,dagplan,llmmap}. BiCICLe coordinates arm-specific LLMs through leader-conditioned trajectories and debate with fresh prompts~\citep{bicicle}. \method\ lets a strong coding agent coordinate persistent member policies as a hyper-agent, revising objectives, arranging synchronization and choosing between delegation and direct lead control of coupled groups. The framework enables concurrency and control transfer while leaving task-dependent coordination strategies to the coding agent.

\section{Asynchronous Coding Agents as Policy}
\label{sec:method}
\method\ organizes several persistent policy agents within one robot. It separates two questions: \emph{how can these agents progress independently, and how should their work be coordinated?} The asynchronous framework enables independent reasoning and execution. A hyper-agent assigns their work and decides when to coordinate actions (\cref{fig:architecture}).

\begin{figure}[!htb]
\centering
\includegraphics[width=0.999\linewidth]{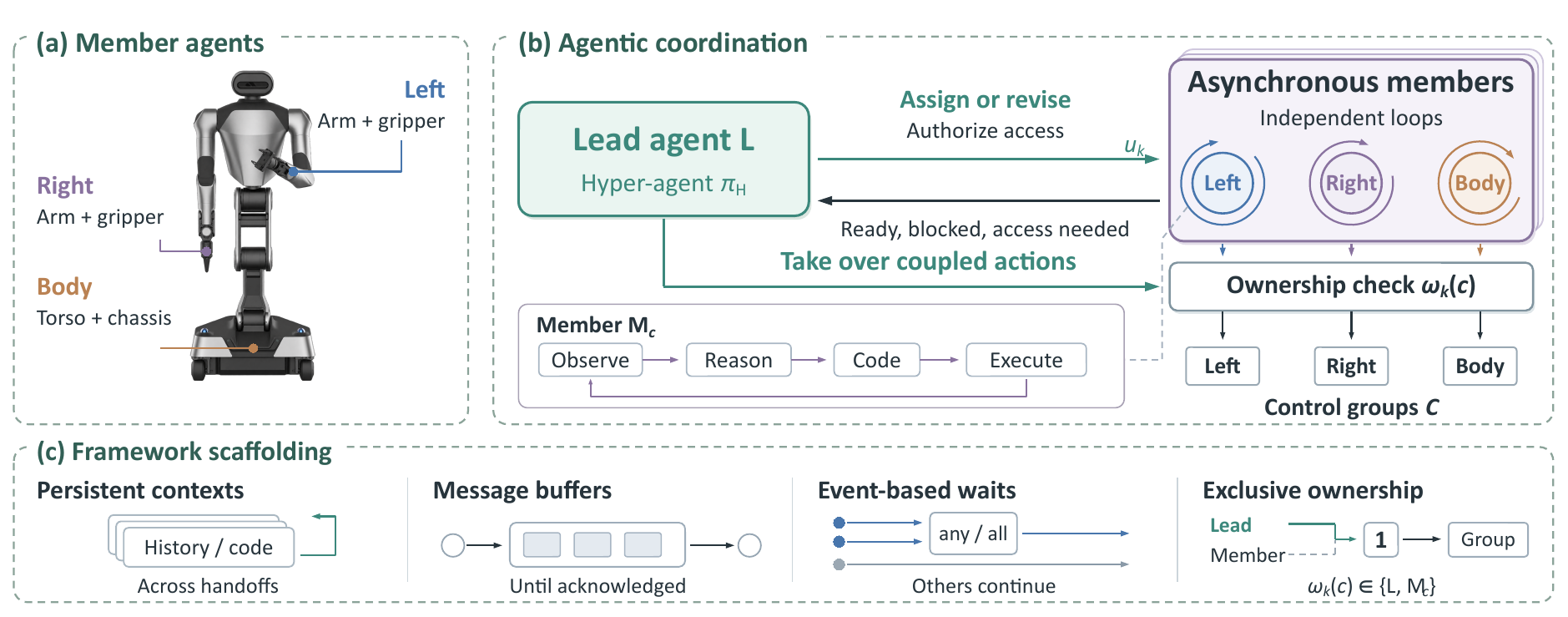}
\caption{\textbf{Independent policy loops with agentic coordination.} (a) Member agents serve configurable control groups, illustrated by two arms with grippers and a body group comprising the torso and chassis. (b) The hyper-agent policy $\pi_H$ uses history $h_k$ and member reports to assign or revise work, authorize access, or arrange direct lead execution of coupled actions. Each member follows an independent observe--reason--code--execute loop. The ownership check $\owner_k(c)$ permits only the current owner to command group $c$. (c) Persistent contexts, message buffers, event-based waits and exclusive ownership support asynchronous progress and coordinated control handoffs.}
\label{fig:architecture}
\end{figure}

\subsection{Asynchronous reasoning and execution}
\label{sec:framework}
Let $\mathcal C$ be the robot's configured control groups, each bundling actuators such as an arm and gripper or a torso and chassis. A persistent member $\member_c$ serves each $c\in\mathcal C$, while a lead $\lead$ can directly control multiple groups. Members are activated as needed.

Each active agent repeatedly observes, reasons and acts at its own pace. One member agent can diagnose a grasp while another plans a placement, without waiting for the other to finish a reasoning turn. Low-level controllers track the commanded motions, allowing both reasoning and physical execution to overlap. Additionally, members can use lighter reasoning for subtask execution while the hyper-agent considers the whole task, keeping the reasoning efficient and within a manageable budget.

\subsection{Agentic coordination with a hyper-agent}
\label{sec:orchestration}
A \emph{hyper-agent} reasons about and coordinates other agents' work. Traditional handcrafted coordination rules encode human assumptions about how work should be organized, which could restrict a capable frontier coding agent's ability to adapt its coordination to the evolving physical situation. \method\ therefore lets the coding agent itself make task-dependent coordination decisions from observations, interaction history and member feedback.

At decision $k$, its policy $\pi_H$ selects an action $u_k$ using the interaction history $h_k$ and current command owners $\owner_k$. The history includes observations, member reports and computed results:
\begin{equation}
 \begin{gathered}
 u_k\sim\pi_H(\,\cdot\mid h_k,\owner_k),\\
 \owner_k(c)\in\{\lead,\member_c\},\quad c\in\mathcal C.
 \end{gathered}
 \label{eq:coordination}
\end{equation}
An action $u_k$ can assign or revise work, request evidence, specify a dependency, authorize access to a shared workspace, or arrange direct lead execution. The index $k$ counts hyper-agent decisions while members follow their own loops. The hyper-agent makes these decisions from feedback rather than following a supplied task graph.

Delegation gives a member an objective, an allowed scope of action and control of its group. Within that scope, it handles local planning and recovery. As dependencies change, the coordinating agent can keep work independent, sequence member actions at a shared workspace, or return control of coupled groups to the lead for a joint action. The hyper-agent can also leave collaborative work with members whose roles complement one another.

\subsection{Framework scaffolding}
\label{sec:synchronization}
The framework provides four mechanisms that support concurrent work and control transfers while leaving task-dependent coordination decisions to the hyper-agent (\cref{fig:architecture}c).

\noindent\textbf{Persistent contexts.} Each member keeps its observations, conversation history and generated programs across assignments and control transfers within an episode. It can continue from previous work without restarting its session or rebuilding its local context. Members can also continue analysis while the lead controls their groups.

\noindent\textbf{Message buffers.} Member agents report to the hyper-agent, which sends objectives, revisions and relevant evidence. Instructions and reports remain in message buffers until the receiving agent confirms receipt. Agents can therefore work at different speeds, with updates kept available while a recipient is busy.

\noindent\textbf{Event-based waits.} A synchronization point specifies \emph{which actions must wait and what evidence permits them to continue}. Agents can wait for any or all specified events while other work continues. The hyper-agent specifies the required events, and the agents check whether the reported physical conditions permit action. For example, during flower insertion in \cref{fig:teaser}, one member prepares and orients its flower while another inserts and withdraws from the vase. The clearance report provides the synchronization event for authorizing the next insertion. Only vase entry waits for the access grant, so preparation can overlap.

\noindent\textbf{Exclusive control ownership.} Command ownership is separate from permission to access a shared workspace. The framework accepts commands for each control group only from its current owner. Ownership cannot change while a call to the robot controller is in progress. Initially, the lead controls each available group. Delegation transfers ownership to a member, and takeover returns it to the lead. Transfers preserve controller targets, allowing responsibility to change without stopping motion or releasing held objects. The new owner checks physical feedback before acting.

\section{Real-world evaluation}
\label{sec:experiments}
\subsection{Task categories}
\label{sec:task_categories}
All experiments use real robots. Following spatial and temporal distinctions in bimanual manipulation~\citep{taxonomy,labor}, we group tasks by dependencies between control groups to test both concurrent work and coordination.

In (i) \emph{Fully Parallel} tasks, control groups execute subtasks independently, as in collecting batteries into separate plates. In (ii) \emph{Parallel-Sequential} tasks, actions depend on another group's progress or access to a shared workspace, as when flowers are prepared independently but inserted into a shared vase in sequence. In (iii) \emph{Collaborative} tasks, control groups jointly support, move or deform an object, as in towel folding.

\subsection{Policy variants}
\label{sec:policy_variants}
All three policies have access to the same observation modalities, numerical tools and robot-control interface:

\noindent\textbf{Direct Policy (Direct).} One coding agent reasons and controls the whole robot. It can issue concurrent robot commands, but planning and control share the same context.

\noindent\textbf{Fixed Delegation (Fixed).} A lead observes, assigns work, replans and synchronizes member agents. Only members command their assigned groups, so assignments and timing can change but ownership cannot return to the lead.

\noindent\textbf{Asynchronous Coding Agents as Policy (ACAP).} The full method in \cref{sec:method}: the coordinating agent chooses assignments and dependencies, retaining delegated execution or returning groups to the lead when direct coordination is appropriate.

\subsection{Evaluation design}
\label{sec:evaluation_design}
We evaluate the efficiency of distributing reasoning and control and the role of agentic coordination in collaborative tasks.

\noindent\textbf{Task-coverage study.} This study evaluates the framework's efficiency and reliability across diverse tasks and robot setups. It comprises 72 episodes across 25 task/platform settings. Direct versus \method\ spans all three dependency categories, while Fixed versus \method\ focuses on the eight Collaborative settings.

\noindent\textbf{Repeated-trial study.} To complement task coverage, this study evaluates variation within each task and the consistency of the observed efficiency gains across executions. It comprises 90 episodes across six tasks, with five trials per task and policy and two tasks per dependency category. We reset the objects to corresponding initial arrangements across policy trials.

The above studies use bimanual manipulation tasks. To examine the framework's applicability to other physical roles, we also extend it to mobile manipulation, with a body member controlling the chassis and torso alongside two arm members (\cref{sec:navigation}).

\subsection{Experimental configuration}
\label{sec:configuration}
We use a Galaxea R1 Pro 2025 robot and four AgileX PiPER-family setups, including PiPER and PiPER-X. Platform details and task assignments are provided in the supplementary material.

The lead agent and coordination hyper-agent share substantial context, and the frontier coding agent is capable of both coordination and direct robot control. We therefore instantiate both roles in a single session, avoiding separate instances with largely overlapping context.
Direct and the lead agents in Fixed and \method\ use Astra with xhigh reasoning. Member agents use Astra with medium reasoning.
Model and reasoning settings remain fixed within an episode. The Direct comparison includes lighter member reasoning. The Fixed comparison holds these settings constant and changes control permissions and role instructions. Each rollout starts with fresh agent sessions, without conversations or task-specific programs from earlier rollouts. Operators monitor safety without providing task guidance or manipulation assistance.

\subsection{Evaluation metrics}
\label{sec:metrics}
\noindent\textbf{Time cost.} Elapsed time $T_i$ is measured from the start of policy execution to the recorded outcome, or to the end of the episode when no outcome is declared. It includes reasoning, computation, waiting, movement and recovery, but excludes task setup before policy execution. In the task-coverage study, the time cost for recorded outcome $o_i$ is
\begin{equation}
L_i=\begin{cases}
T_i,&o_i=\mathrm{completed},\\
\max(T_i,30\text{ min}),&\text{otherwise}.
\end{cases}
\label{eq:cost}
\end{equation}
An unsuccessful early stop incurs at least 30 minutes, while durations beyond 30 minutes remain uncapped.

We average repetitions within each setting, then weight common settings equally. Every episode contributes without giving more weight to settings with more repetitions.

For repeated trials, we report the mean and sample standard deviation of the 30-minute time cost, $L_i^{30}=\min(L_i,30\text{ min})$. Tasks completed within the budget receive their elapsed time, and all other runs receive 30 minutes. Standard deviations across the five trials of each task and policy describe within-task variation.

\noindent\textbf{Physical progress and AUC.} We report graded physical-progress scores on a 0--100 scale, where higher scores indicate greater physical achievement. Human annotators follow a shared task-specific rubric, marking when milestones are first reached and scoring completed work and partial achievement. The rubric defines milestone scores and allows intermediate scores based on the physical quality of the observed result. These annotations define each episode's progress curve, which records the highest score attained by each time. To summarize progress over time, we compute normalized AUC (area under curve) by integrating this curve over 30 minutes and dividing by such a time period. Higher normalized AUC indicates more physical progress achieved earlier. For aggregate comparisons, we average episodes within each setting, then weight settings equally.

\section{Results and analysis}
\label{sec:results}
We first examine efficiency across task dependencies (\cref{sec:asynchrony_results}) and the role of agentic coordination in collaborative manipulation (\cref{sec:coordination_results}). We then assess how these findings vary across repeated executions of the same tasks (\cref{sec:repeated}). Timing analysis examines the associated reasoning, execution and coordination costs (\cref{sec:timing_analysis}). Mobile manipulation examines how the member interface supports different physical roles.

\subsection{Efficiency of the asynchronous framework}
\label{sec:asynchrony_results}
\method\ has a lower task-balanced time cost and higher physical-progress AUC than Direct in each dependency category (\cref{tab:main}). Its advantage extends from independent work to ordered access and coupled manipulation. Even when some actions must be coordinated, members can prepare, perceive and recover independently between those actions. The supplementary material reports physical progress and task completion at five-minute intervals.

\begin{table}[!htb]
\caption{\textbf{Asynchronous framework versus Direct.} Comparison across 25 task/platform settings. Time is task-balanced mean elapsed time in minutes, with a 30-minute minimum for noncompletion. Human annotators assess physical progress. AUC averages physical progress over 30 minutes (0--100), rewarding earlier progress. Bold marks the lowest time and highest AUC.}

\label{tab:main}
\centering
\small
\setlength{\tabcolsep}{6pt}
\begin{tabular}{@{}lrrrr@{}}
\toprule
& \multicolumn{2}{c}{Elapsed time (min) $\downarrow$} & \multicolumn{2}{c}{Progress AUC $\uparrow$} \\
\cmidrule(lr){2-3}\cmidrule(l){4-5}
Task dependency & Direct & \method & Direct & \method \\
\midrule
Fully Parallel (6) & 20.49 & \textbf{13.22} & 60.3 & \textbf{75.2} \\
Parallel-Sequential (11) & 20.27 & \textbf{13.86} & 67.4 & \textbf{73.1} \\
Collaborative (8) & 16.25 & \textbf{13.75} & 56.9 & \textbf{70.8} \\
\midrule
All settings (25) & 19.04 & \textbf{13.67} & 62.3 & \textbf{72.9} \\
\bottomrule
\end{tabular}
\end{table}

\begin{figure}[!htb]
\centering
\includegraphics[width=0.95\linewidth]{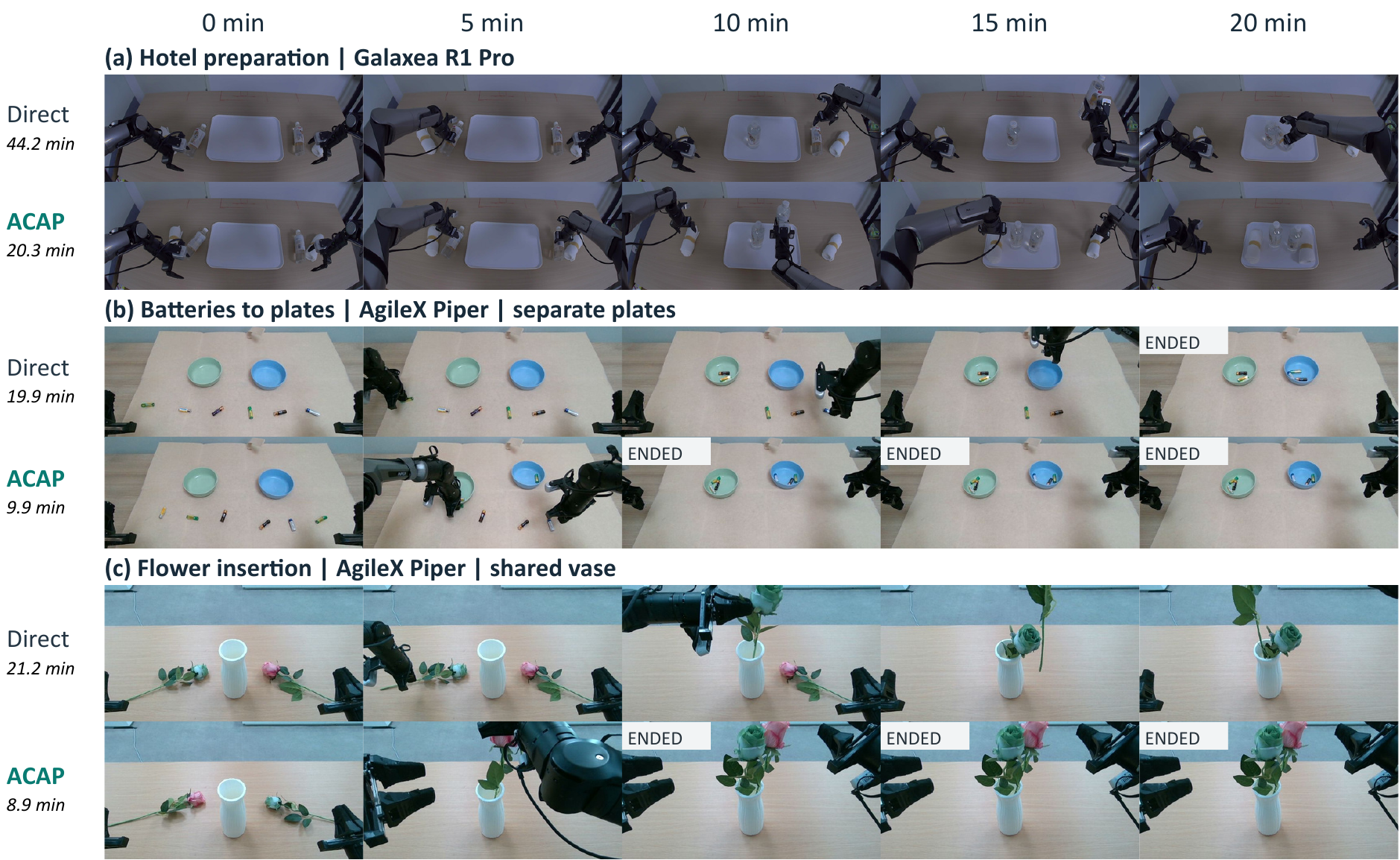}
\caption{\textbf{Visible progress with Direct and ACAP.} Three cases cover Fully Parallel and Parallel-Sequential tasks across Galaxea R1 Pro and two AgileX PiPER installations. Columns share elapsed times within each task. Row labels give observed episode durations, and \textsc{ended} marks repeated final images.}
\label{fig:efficiency}
\end{figure}

\cref{fig:efficiency} connects these differences to physical progress: In battery collection, separate destinations allow independent pickups and placements. In hotel preparation, local pickup and reorientation run in parallel while the lead schedules tray access. In flower insertion, vase entry must be ordered, but stem preparation can proceed alongside insertion (\cref{fig:teaser}). In the two-flower trace, the finished member uses its wrist camera to reveal the other stem's occluded tip. Parallel work remains useful even with shared resources and changing information needs.

\cref{sec:repeated} examines the consistency of these efficiency gains across repeated executions.

\subsection{Agentic coordination for collaborative tasks}
\label{sec:coordination_results}
\method\ has lower mean time cost than Fixed in every Collaborative setting (per-task results in the supplementary material) and improves on Direct on average (\cref{tab:collaborative}). Physical-progress AUC also favors \method\ over both policies on this subset. Fixed already supports multiple agents, lead planning and synchronization. ACAP adds the choice to execute coupled actions directly while retaining delegated work where useful. Its lower average cost supports this flexibility.

\begin{table}[!htb]
\caption{\textbf{Agentic coordination versus fixed delegation.} Results averaged across eight Collaborative settings, reusing their Direct/\method\ episodes from \cref{tab:main}. Fixed and \method\ share agent configurations, with Direct included as a reference. Time, AUC and bold text follow \cref{tab:main}.}

\label{tab:collaborative}
\centering
\small
\setlength{\tabcolsep}{6pt}
\begin{tabular}{@{}lrr@{}}
\toprule
Policy & Elapsed time (min) $\downarrow$ & Progress AUC $\uparrow$ \\
\midrule
Direct & 16.25 & 56.9 \\
Fixed Delegation & 26.02 & 36.3 \\
\method & \textbf{13.75} & \textbf{70.8} \\
\bottomrule
\end{tabular}
\end{table}

\begin{figure}[!htb]
\centering
\includegraphics[width=0.97\linewidth]{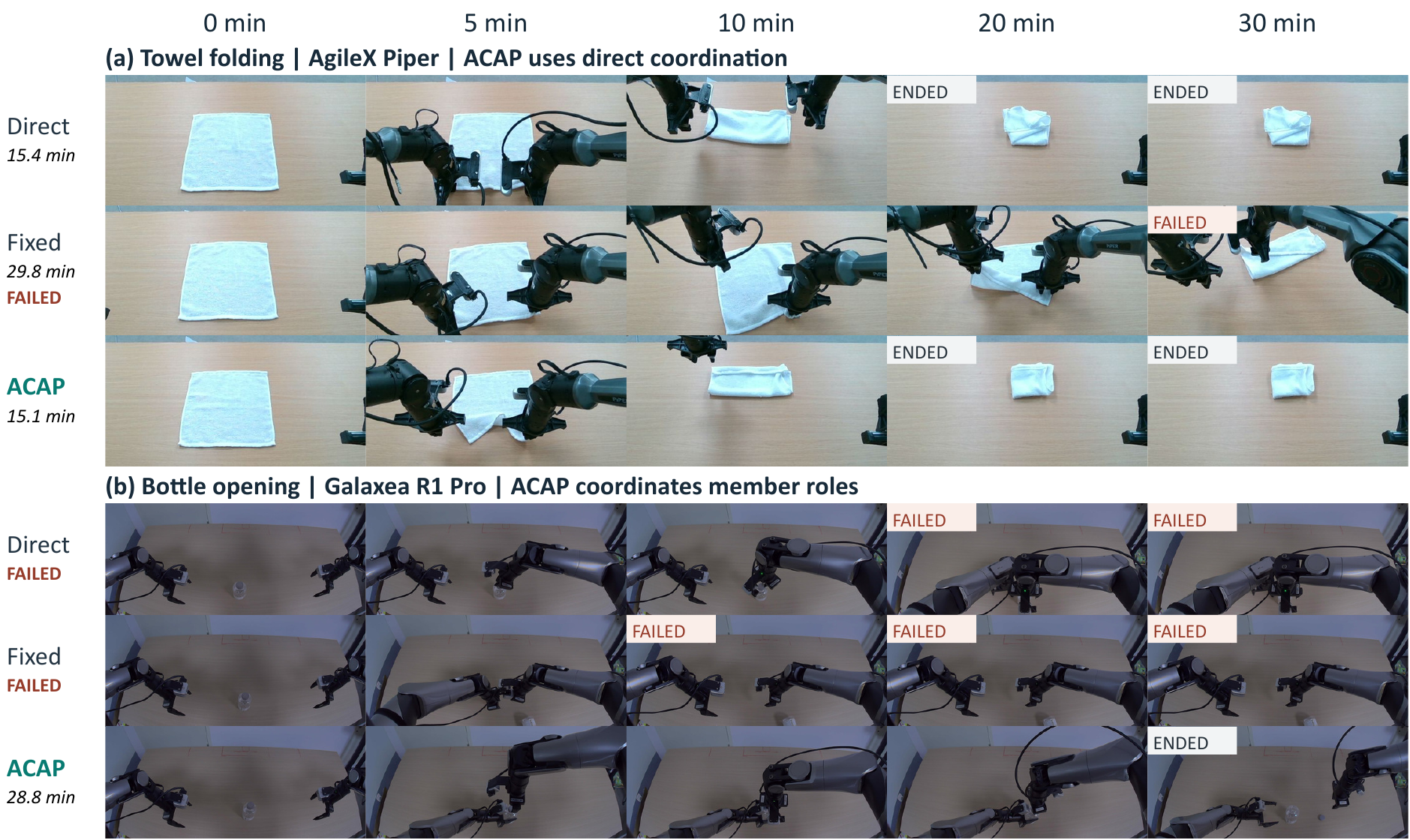}
\caption{\textbf{Coordination on Collaborative tasks.} ACAP uses direct lead execution for towel folding and complementary member roles for bottle opening. Columns share elapsed times. \textsc{Failed} denotes task noncompletion. Both bottle baselines trigger hardware faults after arm crossing. Later Fixed columns repeat its last available image.}
\label{fig:collaboration}
\end{figure}

\cref{fig:collaboration} shows two ways to coordinate a shared object. For towel folding, the lead takes over both arms for a shared fold while retaining member analysis. Completion time is close to Direct, whereas Fixed continues through additional preparation and finishing. For bottle opening, ACAP keeps stabilization and twisting delegated. Only ACAP achieves task completion. Both baselines fail after arm crossing triggers hardware faults. These cases illustrate why agentic coordination includes both organizing member actions and deciding when to act directly.

The Fixed traces show why distributing a shared action can be costly even when it succeeds. In fruit-basket lifting, about 156 seconds elapse between the preparation instruction and the first motion command, including local computation and checks. In carton lifting, one member's wait condition does not recognize the shared start signal, requiring a later corrective command. Both episodes eventually complete, but converting readiness reports into a joint action requires further coordination. The supplementary material analyzes these difficulties and recoveries across the ten Fixed episodes in the task-coverage study.

\subsection{Sensitivity across repeated executions}
\label{sec:repeated}
The repeated trials assess how performance varies within the same task and whether the observed efficiency gains hold consistently across executions. Each of the three policies runs five times on six tasks, with fresh agent sessions for every trial. \cref{tab:repeated} reports time cost and physical-progress AUC within 30 minutes. Individual runs are shown in the supplementary material.

\begin{table}[!htb]
\caption{\textbf{Repeated-trial efficiency and physical progress.} Five trials per task and policy. Time is mean 30-minute-capped cost in minutes, with sample SD in smaller type. AUC uses the physical-progress definition in \cref{tab:main}. Bold marks the lowest mean time and highest AUC. The final row averages the six task means.}
\label{tab:repeated}
\centering
\small
\renewcommand{\arraystretch}{1.22}
\newcommand{\trialcost}[2]{#1\,{\scriptsize\color{black!65}$\pm$\,#2}}
\setlength{\tabcolsep}{1.5pt}
\begin{tabular*}{\linewidth}{@{\extracolsep{\fill}}lrrrrrr@{}}
\toprule
& \multicolumn{2}{c}{Direct} & \multicolumn{2}{c}{Fixed Delegation} & \multicolumn{2}{c}{\method} \\
\cmidrule(lr){2-3}\cmidrule(lr){4-5}\cmidrule(l){6-7}
Task & Time $\downarrow$ & AUC $\uparrow$ & Time $\downarrow$ & AUC $\uparrow$ & Time $\downarrow$ & AUC $\uparrow$ \\
\midrule
\multicolumn{7}{@{}l}{\textbf{Fully Parallel}} \\
Wipe and arrange towels & \trialcost{23.73}{3.84} & 60.5 & \trialcost{15.16}{2.98} & 72.0 & \trialcost{\textbf{12.54}}{1.95} & \textbf{73.7} \\
Sort cubes and batteries & \trialcost{21.61}{3.07} & 62.5 & \trialcost{17.44}{9.45} & 65.8 & \trialcost{\textbf{14.30}}{5.87} & \textbf{75.2} \\
\addlinespace[3pt]
\multicolumn{7}{@{}l}{\textbf{Parallel-Sequential}} \\
Insert four flowers & \trialcost{27.06}{3.79} & 46.7 & \trialcost{25.59}{5.13} & 34.8 & \trialcost{\textbf{22.89}}{6.78} & \textbf{56.6} \\
Assemble a sandwich & \trialcost{22.96}{9.06} & 40.3 & \trialcost{26.50}{7.82} & 42.3 & \trialcost{\textbf{12.17}}{4.70} & \textbf{69.2} \\
\addlinespace[3pt]
\multicolumn{7}{@{}l}{\textbf{Collaborative}} \\
Fold a towel & \trialcost{19.74}{7.98} & 63.6 & \trialcost{27.85}{4.10} & 40.9 & \trialcost{\textbf{16.47}}{3.36} & \textbf{65.1} \\
Lift a milk carton & \trialcost{13.50}{9.34} & \textbf{75.2} & \trialcost{29.82}{0.41} & 35.6 & \trialcost{\textbf{12.45}}{2.37} & 70.1 \\
\midrule
\textbf{All six tasks} & 21.43 & 58.1 & 23.73 & 48.6 & \textbf{15.14} & \textbf{68.3} \\
\bottomrule
\end{tabular*}
\end{table}

\method\ has lower mean capped cost than Direct on all six tasks, with a task-balanced reduction of 29.4\%.

Human assessment also favors \method\ under common physical scoring, with higher physical-progress AUC on five of the six tasks.
The elapsed-time and physical-progress results therefore provide complementary evidence that the efficiency gains persist across repeated executions.

Wiping and towel arrangement gives the clearest separation across trials: every \method\ trial finishes before every Direct trial, and both policies complete all five. This shows a consistent advantage in completing the same physical workload across repeated executions.

The Collaborative trials also favor ACAP over Fixed. \method\ completes ten of ten runs, compared with eight for Direct and three for Fixed. Fixed completes all wiping trials, yet struggles when the arms must jointly fold or support an object. Its ability to distribute independent work therefore does not ensure effective collaboration. Allowing the hyper-agent to revise the division of control remains useful across repeated executions of these tasks.

\subsection{Analysis of efficiency gains}
\label{sec:timing_analysis}
We analyze execution traces to examine how asynchronous work and effective coordination relate to \method's efficiency gains. Reasoning work is the summed duration of agents' active responses. We measure overlaps between agents' reasoning, between reasoning and robot execution, and between control groups' execution. Coordination waiting measures explicit waits. Overlap terms enter negatively in the Direct comparison.

\begin{table}[!htb]
\caption{\textbf{Timing analysis of efficiency gains.} Times are in minutes and exclude outcome penalties. (a) \method\ relative to Direct in both studies. (b) Fixed versus \method\ on repeated Collaborative tasks.}
\label{tab:timing}
\centering
\small
\renewcommand{\arraystretch}{1.12}
\setlength{\tabcolsep}{6pt}
\edef\coordinationrowpadding{\the\dimexpr\arraystretch\baselineskip/3\relax}
\begin{tabular}[t]{@{}lrr@{}}
\multicolumn{3}{@{}l}{\textbf{(a) Asynchrony gains}} \\[3pt]
\toprule
Timing term & Coverage & Repeated \\
\midrule
Extra reasoning work & $+4.95$ & $+6.29$ \\
Reasoning overlap & $-8.17$ & $-9.51$ \\
Reasoning/execution overlap & $-1.35$ & $-1.44$ \\
Execution overlap & $-0.18$ & $-0.14$ \\
\midrule
Elapsed-time difference & $-4.80$ & $-5.23$ \\
\bottomrule
\end{tabular}\hspace{2.5em}%
\begin{tabular}[t]{@{}lrr@{}}
\multicolumn{3}{@{}l}{\textbf{(b) Coordination gains}} \\[3pt]
\toprule
Timing measure & Fixed & \method \\
\midrule
Reasoning work & 46.92 & 23.29 \\[\coordinationrowpadding]
Concurrent reasoning & 14.78 & 6.86 \\[\coordinationrowpadding]
Coordination waiting & 13.86 & 2.85 \\[\coordinationrowpadding]
\midrule
Elapsed time & 28.83 & 14.46 \\
\bottomrule
\end{tabular}
\end{table}

\noindent\textbf{Asynchrony accommodates more agent work in less elapsed time.} We first compare \method\ with Direct to examine how concurrent work relates to elapsed time (\cref{tab:timing}(a)). In both studies, \method\ performs more summed reasoning work but finishes sooner. The timing components show the same pattern across the two studies: overlap between agents' reasoning and between reasoning and robot execution dominates the reported timing gains, while overlap between control groups' execution contributes comparatively little. Thus, the efficiency gain primarily accompanies overlapping reasoning and execution, rather than more simultaneous robot motion alone.

\noindent\textbf{Effective coordination matters beyond enabling parallel agents.} We next examine whether enabling concurrent work is sufficient for efficient collaboration. Fixed and \method\ use the same model configurations and both support asynchronous work, allowing their comparison to examine the role of adaptive coordination. On the two repeated Collaborative tasks, Fixed spends more time reasoning concurrently yet takes longer overall, while \method\ incurs less reasoning work and coordination waiting (\cref{tab:timing}(b)). More concurrent reasoning therefore need not translate into shorter elapsed time. These results support the value of adapting the division of control on coupled tasks, beyond simply allowing agents to run asynchronously.

\subsection{Member agents with different physical roles}
\label{sec:navigation}
We instantiate the general organization in \cref{fig:architecture}(a) with three member agents on a Galaxea R1 Pro: two arms with grippers, and a body member comprising the chassis and torso. The body member uses the same assignment and reporting interface as the arm members. Asked to collect flowers from separate surfaces and place them in a floor basket, the robot travels to the flowers, grasps them and adjusts its body to reach the basket.

Body motion changes the workspace of both arms. The lead waits for carrying-pose reports before authorizing descent, then revises the placement order when one arm cannot reach the basket. The members retain control throughout, while the lead adapts their objectives and dependencies. ACAP completes the task, with both flowers visible in the basket. The supplementary material presents the visual sequence and interactions across the three members.

\section{A dataset of real-world coding-agent rollouts}
\label{sec:dataset}
The rollout dataset links observations, generated programs and numerical results to agent communication, actions, feedback and outcomes. The programs show a recurring process: build a local analytical model, derive an action, and check the object's response. A fitted bottle axis determines a corrective wrist rotation. During carton lifting, a member tracks printed features to correct a mistaken slip diagnosis while the lead controls the robot. These examples connect mathematical reasoning to physical decisions. The supplementary material traces these computations and their revisions and describes the release contents. The dataset makes these processes available for systematic study.

\section{Conclusion}
\method\ organizes one robot as a team of persistent agents assigned to configurable control groups. This definition accommodates different numbers of members and physical roles. The asynchronous framework lets local reasoning and execution progress independently, while a hyper-agent coordinates dependencies and adjusts the division of control. Across the experiments, \method\ reduces mean time cost relative to Direct, and flexible coordination improves on permanent member control for Collaborative tasks. The rollout dataset connects these decisions to generated programs and physical feedback for further study.

\label{maintextend}
\clearpage
\endgroup

\phantomsection
\addcontentsline{toc}{section}{References}
\bibliographystyle{plainnat}
\bibliography{references}
\clearpage
\appendix
\section{Additional quantitative results}
\label[appendix]{app:additional_results}
We compare Direct Policy (Direct), Fixed Delegation (Fixed), and our Asynchronous Coding Agents as Policy (\method).

\subsection{Per-task comparisons}
\label[appendix]{app:per_task_results}
We denote the Galaxea R1 Pro robot as R1 and the four dual-arm PiPER or PiPER-X robots as P-A/B/C/D.

\begingroup
\setlength{\intextsep}{8pt}
\begin{table}[!htbp]
\caption{\textbf{Per-task asynchronous framework comparison.} Direct versus \method\ across the 25 settings in the task-coverage study. We average repetitions within each task. Time is elapsed time in minutes, with a 30-minute minimum for noncompletion. AUC averages human-annotated physical progress over 30 minutes. The lowest time and highest AUC are marked in bold. Detailed task instructions are given in \cref{app:tasks}.}
\label{tab:per_task_framework}
\centering
\small
\setlength{\tabcolsep}{3pt}
\begin{tabular*}{\linewidth}{@{\extracolsep{\fill}}llrrrr@{}}
\toprule
& & \multicolumn{2}{c}{Elapsed time (min) $\downarrow$} & \multicolumn{2}{c}{Progress AUC $\uparrow$} \\
\cmidrule(lr){3-4}\cmidrule(l){5-6}
ID & Task (platform) & Direct & \method & Direct & \method \\
\midrule
\multicolumn{6}{@{}l}{\textbf{Fully Parallel}} \\
T01 & Batteries to plates (P-A) & 18.81 & \textbf{12.31} & 61.6 & \textbf{74.3} \\
T02 & Batteries to boxes (P-C) & 21.26 & \textbf{12.50} & 63.7 & \textbf{77.7} \\
T03 & Wipe and arrange towels (R1) & 15.56 & \textbf{14.46} & 66.8 & \textbf{82.6} \\
T04 & Hotel tray preparation (R1) & 38.17 & \textbf{19.17} & 25.5 & \textbf{56.7} \\
T05 & Towel rolls to tray (R1) & 14.80 & \textbf{11.53} & 70.6 & \textbf{75.9} \\
T06 & Cubes and batteries sorting (P-C) & 14.34 & \textbf{9.35} & 73.4 & \textbf{83.7} \\
\addlinespace[5pt]
\multicolumn{6}{@{}l}{\textbf{Parallel-Sequential}} \\
T07 & Pens to holder (P-A) & 18.06 & \textbf{13.03} & 61.9 & \textbf{70.4} \\
T08 & Flowers to vase (P-B) & 21.25 & \textbf{8.87} & 60.2 & \textbf{86.4} \\
T09 & Grid sorting (P-B) & \textbf{11.37} & 15.21 & \textbf{73.3} & 65.8 \\
T10 & Flowers to vase (P-C) & 55.06 & \textbf{22.53} & 64.3 & \textbf{69.4} \\
T11 & Pens to holder (P-C) & 14.10 & \textbf{11.38} & 74.0 & \textbf{76.9} \\
T12 & Clothes to basket (R1) & 26.53 & \textbf{24.49} & 56.0 & \textbf{56.3} \\
T13 & Sponge cubes to tank (R1) & 13.35 & \textbf{11.25} & 71.9 & \textbf{77.0} \\
T14 & Cups to marked cup (P-B) & 18.55 & \textbf{10.71} & 61.6 & \textbf{71.2} \\
T15 & Toy blocks to basket (P-B) & 18.70 & \textbf{13.62} & 65.3 & \textbf{73.2} \\
T16 & Towel-roll pyramid (P-C) & 15.62 & \textbf{12.17} & 74.6 & \textbf{77.7} \\
T17 & Four flowers to vase (P-C) & 10.34 & \textbf{9.16} & 77.9 & \textbf{80.1} \\
\addlinespace[5pt]
\multicolumn{6}{@{}l}{\textbf{Collaborative}} \\
T18 & Ordered towel fold (P-B) & 14.88 & \textbf{14.24} & 64.0 & \textbf{71.1} \\
T19 & Basket lift (R1) & \textbf{9.94} & 13.01 & \textbf{77.9} & 73.1 \\
T20 & Shirt fold (P-C) & 8.02 & \textbf{6.40} & 73.3 & \textbf{77.4} \\
T21 & Bottle opening (R1) & 30.00 & \textbf{28.76} & 8.5 & \textbf{38.4} \\
T22 & Fruit-basket lift (P-A) & 7.45 & \textbf{6.90} & 80.4 & \textbf{84.9} \\
T23 & Milk-carton lift (P-B) & 21.13 & \textbf{12.28} & 33.5 & \textbf{72.8} \\
T24 & Pot onto book (P-A) & 24.03 & \textbf{15.63} & 47.3 & \textbf{69.4} \\
T25 & Basket onto sponge (P-B) & 14.58 & \textbf{12.78} & 70.1 & \textbf{79.3} \\
\midrule
\multicolumn{2}{@{}l}{All settings (25)} & 19.04 & \textbf{13.67} & 62.3 & \textbf{72.9} \\
\bottomrule
\end{tabular*}
\end{table}
\begin{table}[!ht]
\caption{\textbf{Per-task agentic coordination comparison.} Results on the eight collaborative tasks in \cref{tab:per_task_framework}. Time, AUC and bold text follow \cref{tab:per_task_framework}. The final row averages the eight tasks.}
\label{tab:per_task_coordination}
\centering
\small
\setlength{\tabcolsep}{3pt}
\begin{tabular*}{\linewidth}{@{\extracolsep{\fill}}lrrrrrr@{}}
\toprule
& \multicolumn{3}{c}{Elapsed time (min) $\downarrow$} & \multicolumn{3}{c}{Progress AUC $\uparrow$} \\
\cmidrule(lr){2-4}\cmidrule(l){5-7}
Task & Direct & Fixed & \method & Direct & Fixed & \method \\
\midrule
T18 Ordered towel fold & 14.88 & 34.51 & \textbf{14.24} & 64.0 & 37.4 & \textbf{71.1} \\
T19 Basket lift & \textbf{9.94} & 22.70 & 13.01 & \textbf{77.9} & 46.4 & 73.1 \\
T20 Shirt fold & 8.02 & 8.77 & \textbf{6.40} & 73.3 & 72.1 & \textbf{77.4} \\
T21 Bottle opening & 30.00 & 30.00 & \textbf{28.76} & 8.5 & 9.4 & \textbf{38.4} \\
T22 Fruit-basket lift & 7.45 & 20.59 & \textbf{6.90} & 80.4 & 44.2 & \textbf{84.9} \\
T23 Milk-carton lift & 21.13 & 21.88 & \textbf{12.28} & 33.5 & 46.7 & \textbf{72.8} \\
T24 Pot onto book & 24.03 & 34.85 & \textbf{15.63} & 47.3 & 16.2 & \textbf{69.4} \\
T25 Basket onto sponge & 14.58 & 34.84 & \textbf{12.78} & 70.1 & 18.2 & \textbf{79.3} \\
\midrule
All settings (8) & 16.25 & 26.02 & \textbf{13.75} & 56.9 & 36.3 & \textbf{70.8} \\
\bottomrule
\end{tabular*}
\end{table}
\endgroup

\subsection{Physical progress and task completion over time}
\label[appendix]{app:progress}
In \cref{tab:budgets}, we report physical progress and task completion at five-minute intervals up to 30 minutes, using the settings in \cref{tab:per_task_framework}. Episodes are averaged within each task, and tasks are then weighted equally for overall results. Human annotators follow task-specific rubrics, marking when milestones are first reached and scoring completed work and partial achievement. We show the annotation interface in \cref{app:annotation}.

\begin{table}[!htbp]
\caption{\textbf{Physical progress and task completion at five-minute intervals.} $P_t$ is physical progress by time $t$ (0--100). $C_t$ is task completion by time $t$ (\%). Episodes are averaged within each task, then averaged equally across tasks. Bold marks the best within each group.}
\label{tab:budgets}
\centering
\begin{tabular*}{\linewidth}{@{\extracolsep{\fill}}llrrrrrr@{}}
\toprule
Policy & Measure & 5 min & 10 min & 15 min & 20 min & 25 min & 30 min \\
\midrule
\multicolumn{8}{@{}l}{\textbf{Fully Parallel (6 settings)}} \\
\multirow{2}{*}{Direct} & Progress $P_t$ & 14.6 & 49.3 & 76.4 & 87.5 & 91.7 & 93.8 \\
 & Completion $C_t$ & 0.0 & 0.0 & 27.8 & 66.7 & 83.3 & 83.3 \\
\addlinespace[3pt]
\multirow{2}{*}{\method} & Progress $P_t$ & \textbf{29.2} & \textbf{81.3} & \textbf{93.8} & \textbf{100.0} & \textbf{100.0} & \textbf{100.0} \\
 & Completion $C_t$ & 0.0 & \textbf{25.0} & \textbf{83.3} & \textbf{91.7} & \textbf{100.0} & \textbf{100.0} \\
\addlinespace[6pt]
\multicolumn{8}{@{}l}{\textbf{Parallel-Sequential (11 settings)}} \\
\multirow{2}{*}{Direct} & Progress $P_t$ & 14.8 & 61.8 & 86.0 & \textbf{96.9} & 98.9 & 98.9 \\
 & Completion $C_t$ & 0.0 & 0.0 & 36.4 & 72.7 & 81.8 & 90.9 \\
\addlinespace[3pt]
\multirow{2}{*}{\method} & Progress $P_t$ & \textbf{28.4} & \textbf{71.6} & \textbf{93.2} & 96.1 & \textbf{99.1} & \textbf{99.1} \\
 & Completion $C_t$ & 0.0 & \textbf{18.2} & \textbf{72.7} & \textbf{81.8} & \textbf{100.0} & \textbf{100.0} \\
\addlinespace[6pt]
\multicolumn{8}{@{}l}{\textbf{Collaborative (8 settings)}} \\
\multirow{2}{*}{Direct} & Progress $P_t$ & 25.0 & 56.9 & 65.6 & 71.9 & 84.4 & 84.4 \\
 & Completion $C_t$ & 0.0 & \textbf{31.2} & 56.2 & 62.5 & \textbf{87.5} & 87.5 \\
\addlinespace[3pt]
\multirow{2}{*}{Fixed Delegation} & Progress $P_t$ & 11.6 & 26.9 & 39.4 & 53.8 & 58.4 & 59.1 \\
 & Completion $C_t$ & 0.0 & 12.5 & 12.5 & 18.8 & 43.8 & 43.8 \\
\addlinespace[3pt]
\multirow{2}{*}{\method} & Progress $P_t$ & \textbf{29.4} & \textbf{69.7} & \textbf{89.4} & \textbf{92.5} & \textbf{95.6} & \textbf{98.8} \\
 & Completion $C_t$ & 0.0 & 25.0 & \textbf{68.8} & \textbf{87.5} & \textbf{87.5} & \textbf{100.0} \\
\addlinespace[6pt]\midrule
\multicolumn{8}{@{}l}{\textbf{All 25 settings}} \\
\multirow{2}{*}{Direct} & Progress $P_t$ & 18.0 & 57.2 & 77.2 & 86.6 & 92.5 & 93.0 \\
 & Completion $C_t$ & 0.0 & 10.0 & 40.7 & 68.0 & 84.0 & 88.0 \\
\addlinespace[3pt]
\multirow{2}{*}{\method} & Progress $P_t$ & \textbf{28.9} & \textbf{73.3} & \textbf{92.1} & \textbf{95.9} & \textbf{98.2} & \textbf{99.2} \\
 & Completion $C_t$ & 0.0 & \textbf{22.0} & \textbf{74.0} & \textbf{86.0} & \textbf{96.0} & \textbf{100.0} \\
\bottomrule
\end{tabular*}
\end{table}

Across all settings, \method\ achieves a higher task completion rate at each reported time from ten minutes onward. Lower average time cost also demonstrates earlier overall task completion.

Physical progress reveals this advantage earlier in execution: \method\ is ahead of Direct at five minutes, before either policy completes a task, and remains ahead at every reported time in the overall comparison. On the collaborative tasks, \method\ also has the highest progress score at each time.

\subsection{Variation across repeated executions}
\begin{figure}[!htbp]
\centering
\includegraphics[width=0.99\linewidth]{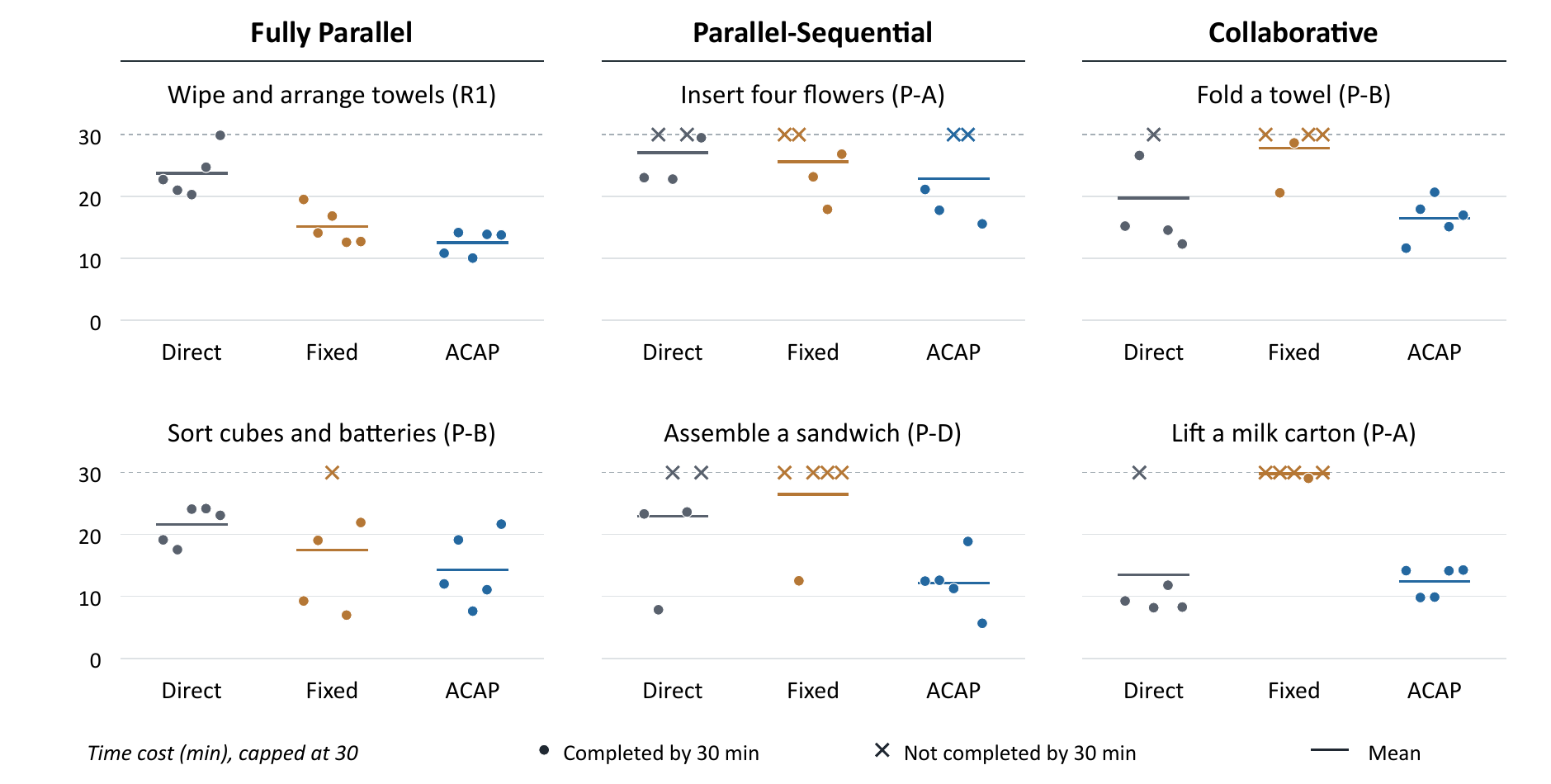}
\caption{\textbf{Time cost across repeated executions.} Dots denote task completion within 30 minutes, and crosses denote incomplete tasks at 30 minutes. Horizontal bars show means. Each task and policy has five trials.}
\label{fig:repeated_distribution}
\end{figure}

\label[appendix]{app:repeated_variation}
In \cref{fig:repeated_distribution}, we visualize the sensitivity results from the main paper, showing how time cost and task completion vary across repeated executions of the same task.

Two patterns are visible across the trial distributions. In tasks that can be fully or partly parallelized, \method\ shows clear efficiency benefits from asynchronous reasoning and execution. In tasks requiring coupled actions, agentic coordination enables \method\ to preserve the reliability and time efficiency of Direct while improving on Fixed. The figure therefore makes both aspects of the result visible: faster execution of independent work and more reliable completion of coupled actions.

\section{Task definitions and evaluation protocols}
\label[appendix]{app:fullresults}
\subsection{Task-coverage tasks and protocol}
\label[appendix]{app:tasks}
\cref{tab:instructions} lists the instructions for the 25 task/platform settings in the task-coverage study. Task IDs match the per-task results in \cref{tab:per_task_framework,tab:per_task_coordination}.

\begin{table}[!htbp]
\caption{Task instructions for all 25 settings, grouped by dependency category.}
\label{tab:instructions}
\centering
\begin{tabular}{@{}l>{\raggedright\arraybackslash}p{\dimexpr\linewidth-3.2em\relax}@{}}
\toprule ID & Instruction \\
\midrule
\inputrows sections/tables/task_instructions.tex
\bottomrule
\end{tabular}
\end{table}

\subsection{Repeated-trial tasks and protocol}
\label[appendix]{app:repeated_protocol}
\cref{tab:repeated_instructions} shows the task instructions and robot installations for the six repeated-trial tasks. Task instructions, robot installation, model, reasoning settings and body configuration remain constant across repetitions within each task and policy.

\begin{table}[!htbp]
\caption{Instructions in the repeated-trial study. The same instruction is used for all five trials of each policy on a task.}
\label{tab:repeated_instructions}
\centering
\begin{tabularx}{\linewidth}{@{}>{\raggedright\arraybackslash}p{0.28\linewidth}Y@{}}
\toprule Task (platform) & Instruction \\
\midrule
\inputrows sections/tables/repeated_instructions.tex
\bottomrule
\end{tabularx}
\end{table}
\FloatBarrier

\subsection{Physical-progress annotation}
\begin{figure}[!htbp]
\centering
\includegraphics[width=\linewidth]{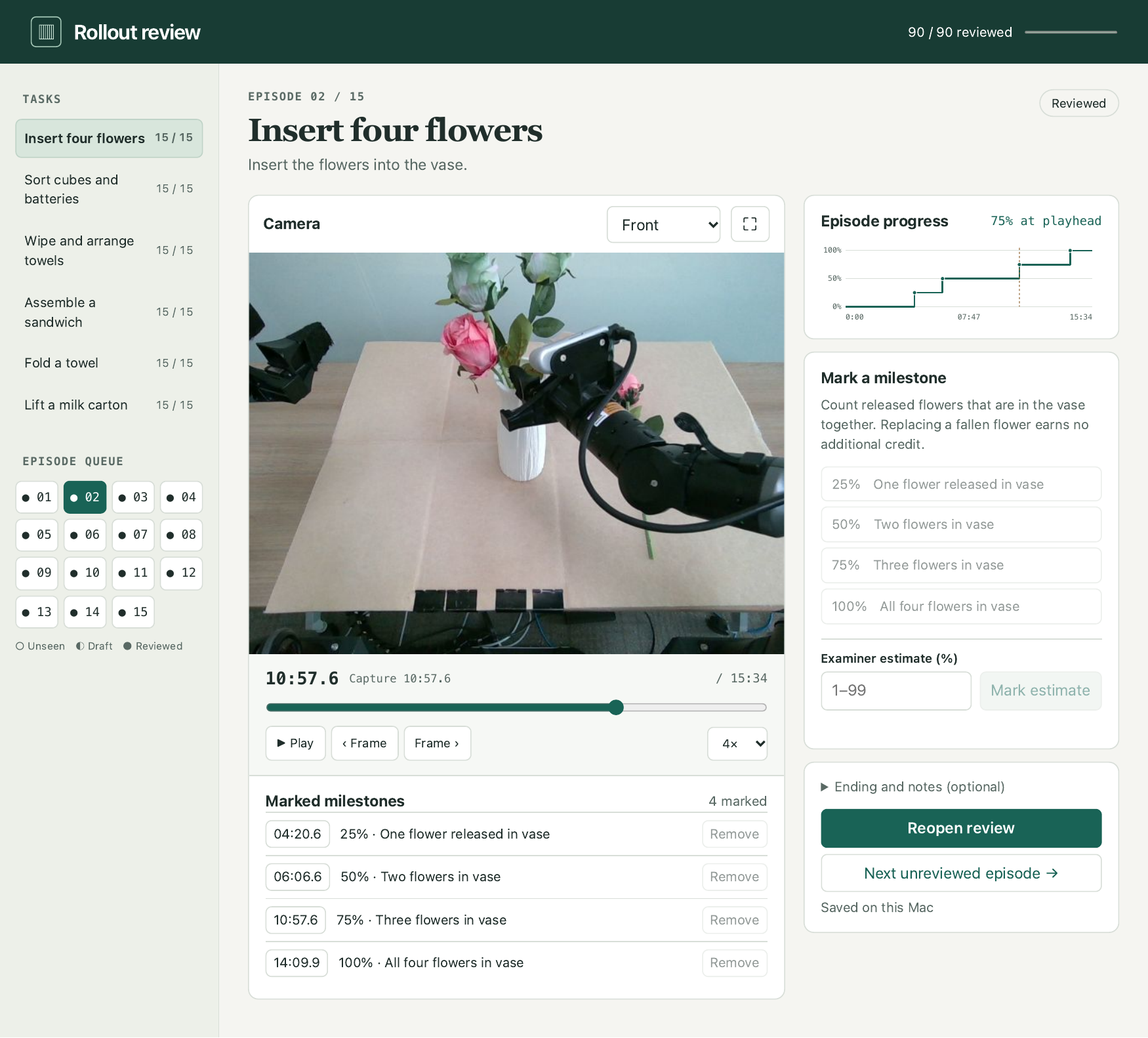}
\caption{\textbf{Physical-progress annotation interface.} Camera frames, task-specific scoring criteria and timestamped marks are shown together with the episode's progress curve. The selected frame shows three inserted flowers, corresponding to 75\% progress.}
\label{fig:annotation_interface}
\end{figure}

\label[appendix]{app:annotation}
\Cref{fig:annotation_interface} shows the human annotation interface for reviewing physical progress. It presents the task instruction and camera recordings alongside task-specific scoring criteria, with policy labels hidden. Annotators can switch views and step through original frames to mark when a milestone is first achieved. Each mark stores a progress score, elapsed time and the corresponding frame. The milestone list at the bottom and progress curve on the right allow annotators to revisit and check these judgments.

We provide task-specific rubrics defining the main milestones and their scores. For example, four-flower insertion assigns 25\% per flower released in the vase, while towel folding uses the first completed fold as an intermediate milestone. Annotators can also assign intermediate scores based on the physical quality of the observed result and add notes explaining their judgments. These timestamped annotations form the progress curves summarized by AUC.

\section{Coordination costs under fixed delegation}
\label[appendix]{app:coordination_analysis}
Coupled manipulation makes the members' plans interdependent. In fruit-basket lifting, one arm's grasp attempt moves the basket, shifting the handle the other arm is approaching. This prompts another round of communication and planning to update the grasp target and coordinate the next attempt. The members must also synchronize their separately prepared motion programs. These updates and checks add extra reasoning and waiting before the arms can act together.

Fixed and \method\ use the same model configurations and both support asynchronous work. The timing analysis in the main paper shows that Fixed spends more time reasoning concurrently yet takes longer overall, while \method\ incurs less reasoning work and coordination waiting. Therefore, overlapping additional reasoning rounds can coexist with delayed physical progress when the members repeatedly revise and synchronize their actions.

\method\ lets the hyper-agent decide when to keep work delegated and when to bring coupled actions under direct lead control. In the towel-folding case from the main paper, members establish their grasps before the lead controls both arms for the shared fold. Direct control allows the lead to coordinate the following coupled motion without repeatedly exchanging instructions and reports between separate member programs. Choosing when to take over addresses this coordination overhead while retaining asynchronous preparation and local reasoning where they are useful.

\section{Mobile manipulation with a body member}
\begin{figure}[!htbp]
\centering
\includegraphics[width=\linewidth]{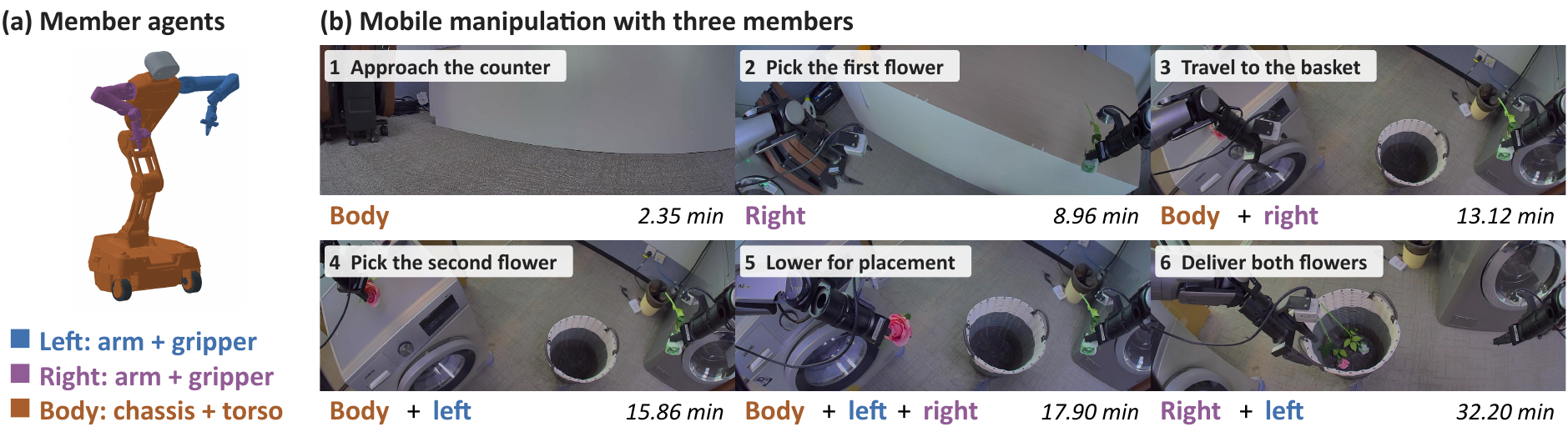}
\caption{\textbf{Coordinating body and arm members.} (a) An instantiation with three control groups, colored by member: left arm and gripper, right arm and gripper, and chassis and torso. (b) Combined member labels indicate complementary roles: body navigation with a retained right-hand grasp (3), torso planning during left-member acquisition (4), and body lowering while both arm members maintain their grasps (5). Images come from chassis- and head-mounted cameras.}
\label{fig:navigation}
\end{figure}

\label[appendix]{app:navigation}
We extend the same asynchronous framework and agentic coordination to members with different physical responsibilities. On a Galaxea R1 Pro robot, we instantiate a body agent controlling the chassis and torso alongside the two arm agents. The body member uses the same assignment and reporting interface as the arms, with responsibility for navigation and torso positioning. The robot is asked to collect flowers from separate locations and put them in a floor-level basket. \cref{fig:navigation} illustrates the member roles and task progression.

\paragraph{Asynchronous reasoning and execution.}
Independent work continues across the body and arm responsibilities. For example, the body agent plans chassis realignment and the right agent evaluates flower-placement poses while the left arm retracts with its flower. Elsewhere, the body executes a chassis adjustment while the left agent interprets new views of the destination basket. These interactions show navigation planning, scene interpretation and arm execution participating in the same asynchronous organization. \cref{tab:navigation_events} gives the corresponding intervals and coordination events.

\paragraph{Agentic coordination across body and arm members.}
Body motion changes the workspace of both arms, introducing dependencies between navigation and manipulation. Before lowering the torso, the hyper-agent waits for both arm members to report that their flowers are retained and their arms are ready for the movement. Later, when the left member reports that it cannot reach the basket from the current alignment, the hyper-agent revises the order: the right member places its flower first, the body realigns the robot, and the left member then places its flower. The hyper-agent uses member reports to decide when each stage can proceed. Throughout these changes, the members retain control of their groups while the hyper-agent adapts their objectives and dependencies. Both asynchronous progress and agentic coordination therefore carry over to the extended member definition.

\begin{table}[!htbp]
\caption{\textbf{Asynchronous work and coordination in mobile manipulation.} Selected interactions from the body-agent rollout. Times are elapsed mm:ss from episode start.}
\label{tab:navigation_events}
\small
\renewcommand{\arraystretch}{1.12}
\begin{tabularx}{\linewidth}{@{}>{\raggedright\arraybackslash}p{0.26\linewidth}Y@{}}
\toprule
Interaction & Timestamped evidence \\
\midrule
Route planning during arm execution & 08:45--09:13: The body agent plans the navigation route. Meanwhile, the right arm retracts with its flower during 08:49--08:58 and reports readiness for transport at 09:11. \\
\addlinespace[5pt]
Chassis movement during scene interpretation & 09:37--09:55: The left agent reassesses the destination basket from newly available views. During this assessment, the body executes a backward chassis movement at 09:48--09:51. The left agent's report subsequently informs the navigation direction. \\
\addlinespace[5pt]
Synchronization before torso movement & The right agent reports readiness at 15:39, and the left reports its flower retained after inward retraction at 16:48. After both reports, the lead authorizes the body agent to lower the torso at 17:05, with the arms holding their positions. \\
\addlinespace[5pt]
Concurrent work across all three members & 20:46--21:02: The body agent plans chassis realignment, the right agent evaluates flower-placement poses, and the left arm physically retracts with its flower. All three members progress concurrently. \\
\addlinespace[5pt]
Body realignment followed by arm placement & The right agent reports completed placement and a stable empty arm at 27:02. The lead authorizes chassis realignment at 27:10. After the body reports completion at 28:36, the lead authorizes left-arm placement at 28:50, which the left agent reports completed at 32:02. \\
\bottomrule
\end{tabularx}
\end{table}

\section{Rollout dataset for studying embodied coding agents}
\label[appendix]{app:implementation}
Our dataset contains 162 real-world rollouts from Direct, Fixed and \method. It preserves how coding agents construct partial geometric representations, reason within them and translate the resulting plans into robot actions. The examples below show how the saved programs, observations and execution records make these decisions available for study.

\subsection{Dataset contents}
The dataset contains task instructions, timestamped observations, generated programs, numerical outputs, action requests, robot feedback and task outcomes. Programs saved in \texttt{development/} and agent workspaces preserve the computations performed during execution. Available RGB and depth observations, calibration and measured robot state provide their physical context. Observation identifiers and timestamps allow a computation to be traced from its input observations to the issued end-effector or joint targets, gripper commands and subsequent feedback.

Multi-agent rollouts also retain assignments, messages and control transfers, showing how agents request, share and use each other's analysis. An index supports retrieval by task, platform, policy and task-dependency category. The framework and role instructions accompany the dataset.

\subsection{Representing, planning and acting through code}
\begin{figure}[!htbp]
\centering
\includegraphics[width=\linewidth]{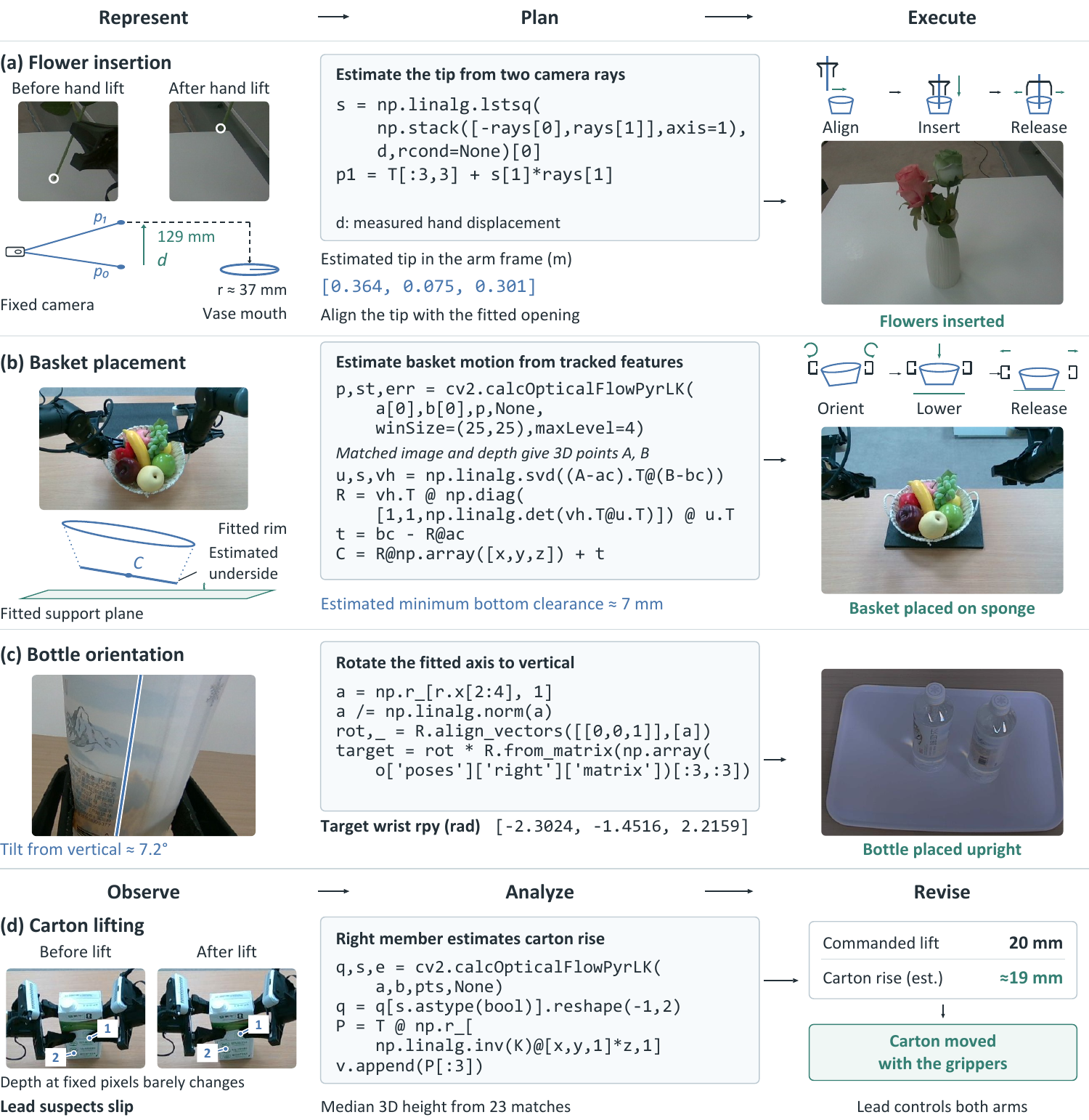}
\caption{\textbf{Geometric reasoning for action and feedback.} (a) The robot raises the flower. The agent uses the measured hand displacement and two images to estimate the tip's position, then aligns and inserts the stem. (b) The agent estimates basket placement and clearance from fitted surfaces and tracked motion, then adjusts both arms to lower and release the basket. (c) A fitted bottle axis determines a corrective wrist rotation, which is used in the next motion command before tray placement. (d) After a commanded 20\,mm lift, tracked carton features indicate a rise of approximately 19\,mm, correcting the lead's initial slip diagnosis.}
\label{fig:dataset_reasoning}
\end{figure}

\label[appendix]{app:geometric_models}
\Cref{fig:dataset_reasoning} illustrates how agents represent a scene, plan within that representation, issue robot commands and check the resulting motion. The programs are generated during execution using observations, calibration and robot interfaces.

\paragraph{Locating a flower tip and planning its insertion.}
The agent represents the vase opening as a circle fitted to depth points and the thin stem tip as an unknown point along an image ray. To locate its flower's tip, the right member raises its hand while keeping the grasp orientation approximately fixed. The tip's image positions before and after this movement define two rays from the same camera. The measured hand translation constrains the displacement between the corresponding tip positions. A generated NumPy program solves for their depths by least squares, yielding an approximate three-dimensional tip position without relying on depth measurements, which could be unreliable here because the stem is thin.

The agent can then reason about the tip relative to the opening: the approach should bring the tip above the rim before descent. The member moves its hand above the vase, refines the tip-to-rim alignment from a closer observation, and commands incremental downward motions. It then opens the gripper and withdraws once the stem is contained. In this example, the model turns an image-localization problem into a spatial alignment and insertion plan that the robot executes.

\paragraph{Modeling basket placement and support clearance.}
To put a fruit basket on a sponge, the lead fits support planes from depth points and a circle to the basket rim. These models provide the sponge's target region and the basket's center, turning placement into a displacement between geometric locations. The agent computes the horizontal translation and carries the basket with paired arm movements.

As the basket moves, the agent also estimates where its underside, which is invisible from all available cameras, lies relative to the sponge. Generated OpenCV code tracks basket features across observations and backprojects the matches into three dimensions. Singular value decomposition (SVD) then fits a rigid transformation between these matched points. Applying this transformation to an initial model of the basket bottom yields estimates of its position, tilt and clearance above the support plane.

The agent uses this geometry with fresh observations to adjust both wrist orientations and lower the basket incrementally. It then opens both grippers for basket placement and withdraws. The final images show the basket resting on the sponge with the fruit retained. This example shows how a compact scene model and motion analysis guide coordinated wrist, arm and gripper actions.

\paragraph{Estimating a bottle axis to correct its orientation.}
In hotel tray preparation, the right member backprojects wrist-camera depth points into the robot base frame and fits a cylinder using robust least squares. It computes the rotation that aligns the fitted bottle axis with the vertical direction and applies it to the current end-effector orientation. The next motion command uses this orientation to correct the bottle's tilt.

\paragraph{Revising an interpretation of physical feedback.}
After commanding the grippers to lift the carton, the lead observes nearly unchanged depth at fixed image locations. It initially interprets this as the carton remaining in place while the grippers move upward, suggesting that they have not grasped it securely. While the lead controls both arms, the right member writes OpenCV code to track printed carton features and backproject the matched depth measurements into three dimensions. Following the same physical points reveals that the carton rises with the grippers. The lead accepts this evidence and revises its slip diagnosis. The code, numerical results and messages preserve how geometric evidence changes the lead's interpretation of execution feedback.

\section{Limitations and future work}
\label{sec:discussion}
Our rollouts show agents reasoning about potential interference, sequencing access to shared workspaces, checking clearance from observations and revising motion plans. These behaviors demonstrate collision awareness at the policy level. The current framework does not include a dedicated motion-level collision-checking layer. A complementary control layer could incorporate full robot geometry, including mounted cameras and other attachments, together with environment geometry to check planned motions and monitor execution. Developing and evaluating this extension is left to future work.

\section{Additional visual comparisons}
\begin{figure}[!htp]
\centering
\includegraphics[width=\linewidth,height=540bp,keepaspectratio]{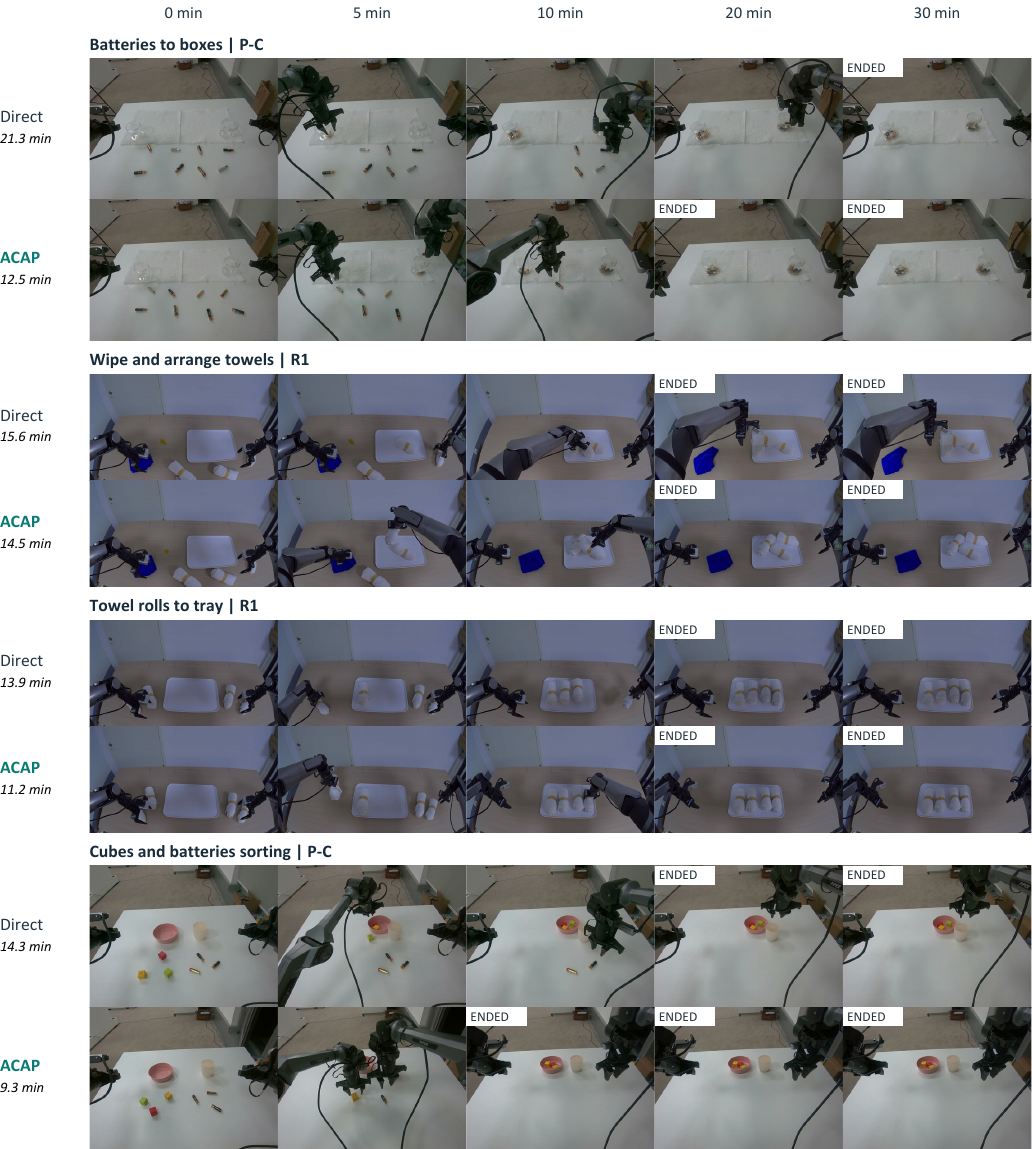}
\caption{\textbf{Fully Parallel tasks.} Direct and \method\ perform battery placement, wiping and towel arrangement, towel-roll placement, and independent cube/battery sorting.}
\label{fig:supp1}
\end{figure}

\begin{figure}[!p]
\centering
\includegraphics[width=\linewidth,height=590bp,keepaspectratio]{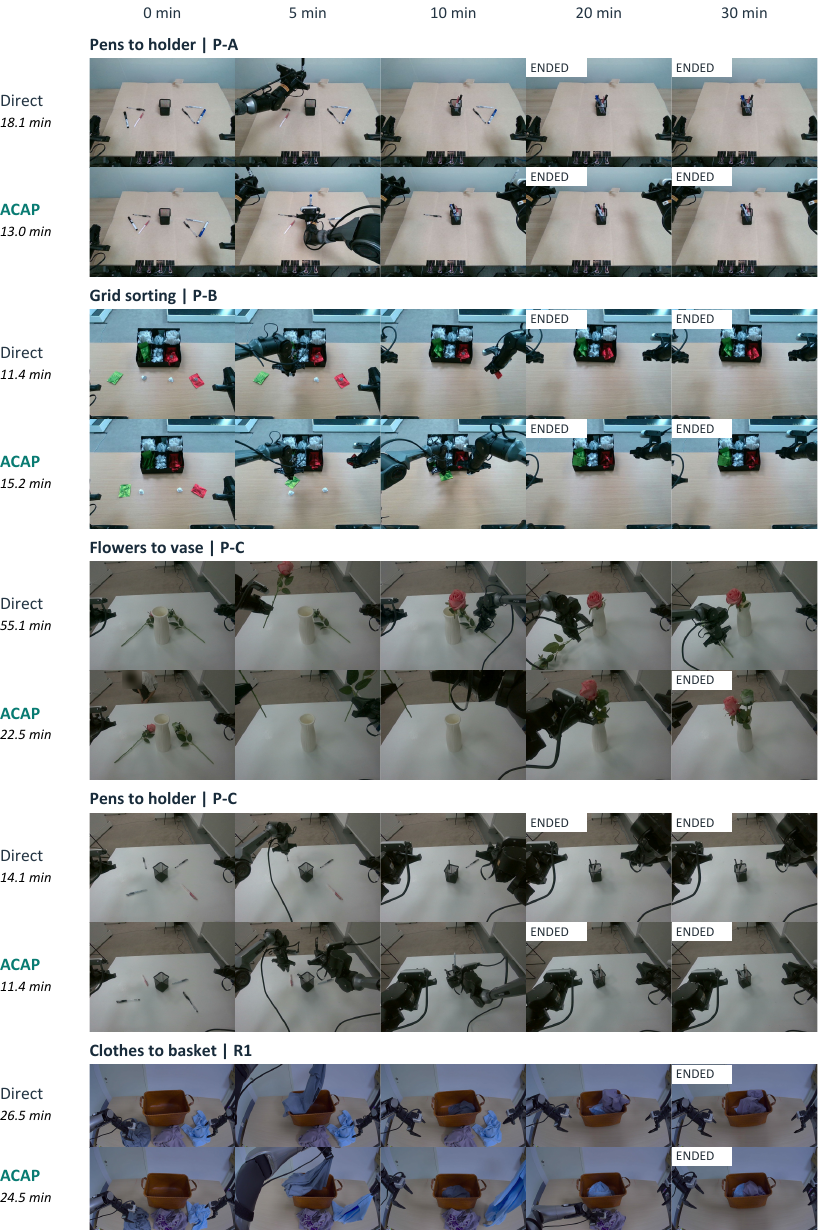}
\caption{\textbf{Parallel-Sequential tasks.} Direct and \method\ perform pen placement on two installations, grid sorting, flower insertion and clothes collection.}
\label{fig:supp_parallel_sequential_1}
\end{figure}

\begin{figure}[!p]
\centering
\includegraphics[width=\linewidth,height=590bp,keepaspectratio]{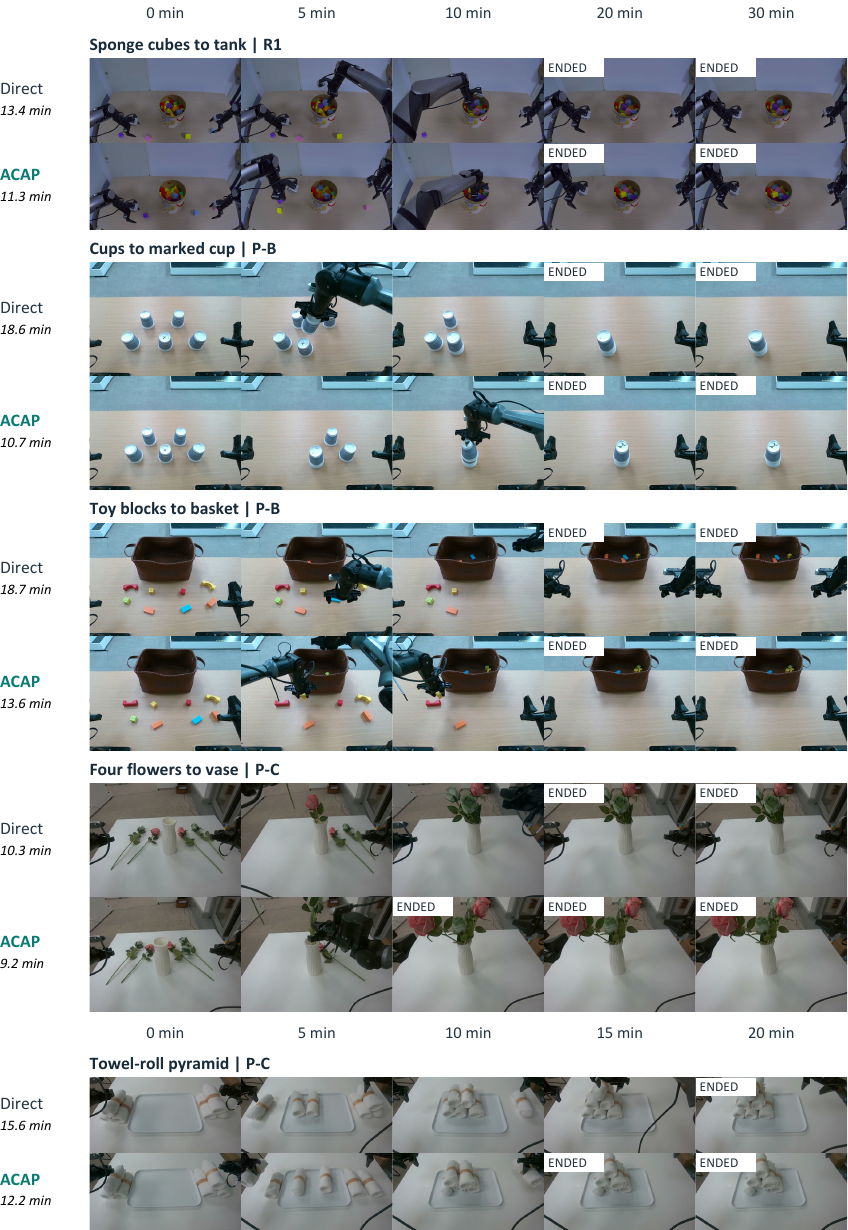}
\caption{\textbf{Parallel-Sequential tasks (continued).} Sponge-cube placement, cup stacking, toy-block collection, four-flower insertion and towel-pyramid construction. The towel-pyramid sequence has its own column times.}
\label{fig:supp_parallel_sequential_2}
\end{figure}

\begin{figure}[!p]
\centering
\includegraphics[width=\linewidth,height=590bp,keepaspectratio]{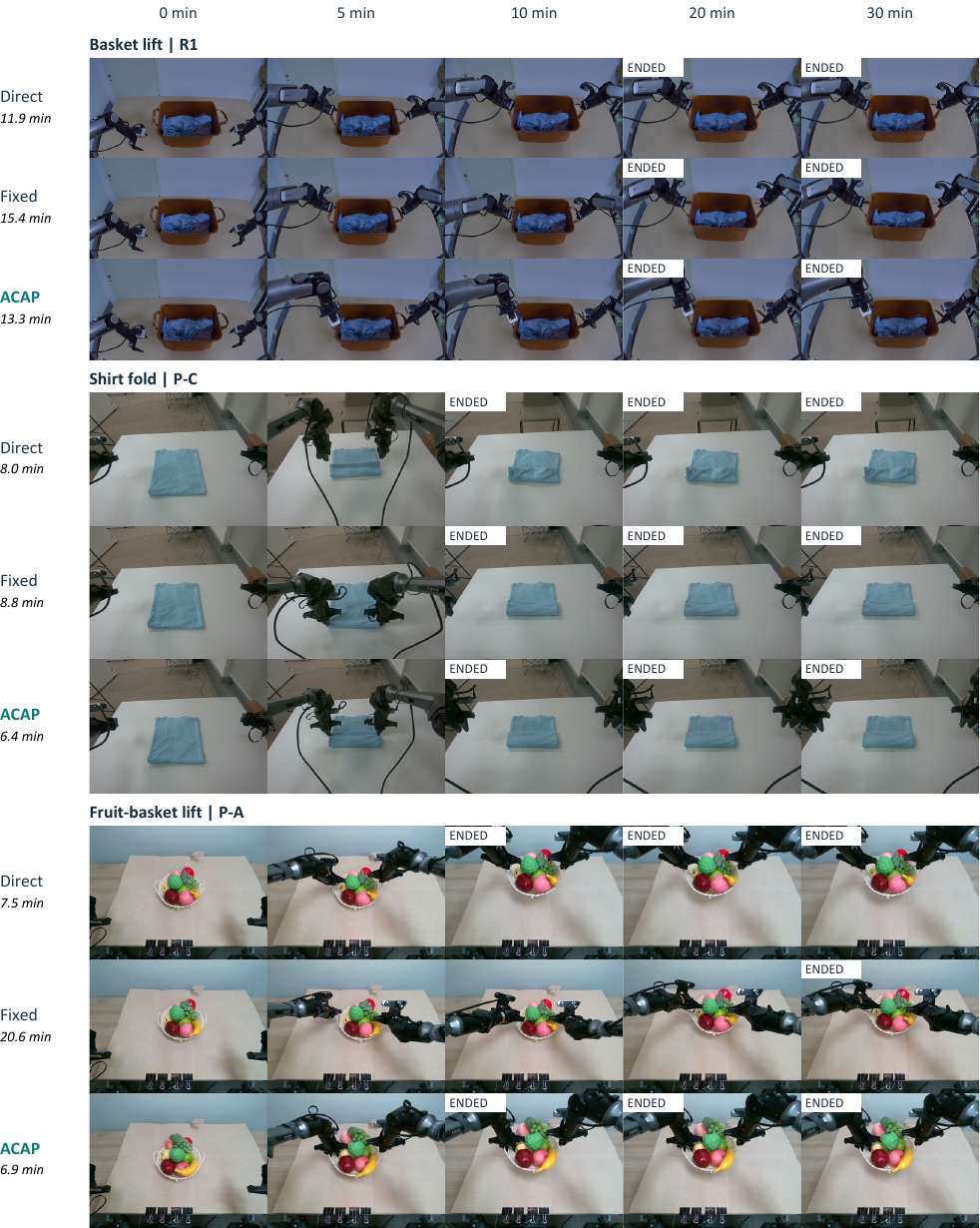}
\caption{\textbf{Collaborative tasks.} Direct, Fixed and \method\ perform basket lifting on R1, shirt folding on P-C and fruit-basket lifting on P-A.}
\label{fig:supp_collaborative_1}
\end{figure}

\begin{figure}[!p]
\centering
\includegraphics[width=\linewidth,height=590bp,keepaspectratio]{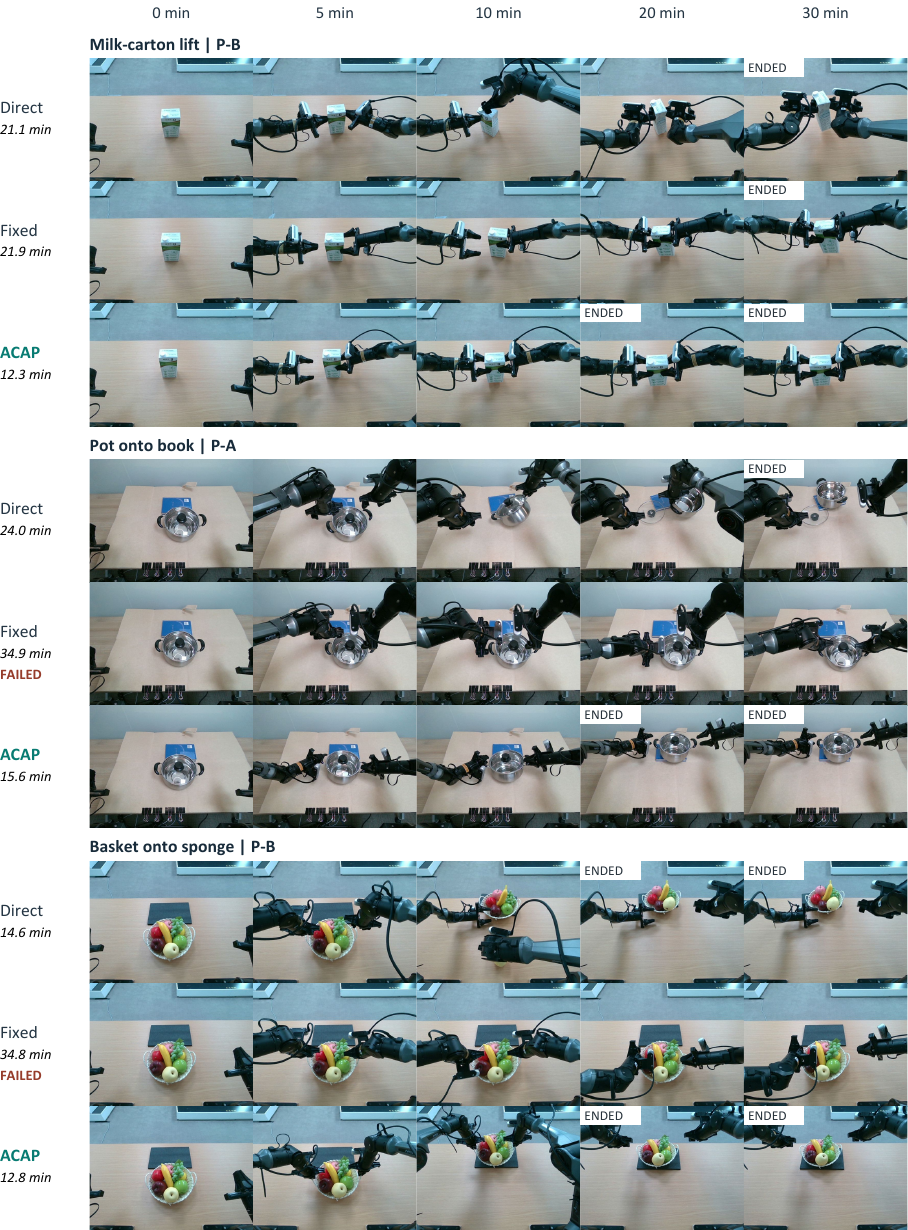}
\caption{\textbf{Collaborative tasks (continued).} Milk-carton lifting, pot placement on a book and fruit-basket placement on a sponge. Direct and \method\ achieve task completion in all three settings. Fixed leaves the pot and basket tasks unfinished in these runs.}
\label{fig:supp_collaborative_2}
\end{figure}

\label[appendix]{app:cases}
\crefrange{fig:supp1}{fig:supp_collaborative_2} provide additional policy comparisons. Rows are policies and columns are elapsed times. \textsc{ended} marks repeated final images, and \textsc{failed} indicates an unsuccessful outcome.

\end{document}